\documentclass[acmlarge]{acmart}

\usepackage{mathtools} 

\usepackage{booktabs}
\usepackage{xltabular}
\usepackage{multirow}

\usepackage{algorithm}
\usepackage{algpseudocodex}

\usepackage[inline]{enumitem} 
\usepackage{amsthm} 
\usepackage{amsthm} 
\usepackage{wrapfig}

\theoremstyle{definition}
\newtheorem{definition}{Definition}[]

\usepackage{caption}
\usepackage{subcaption}

\usepackage{tikz}
\usepackage[subpreambles=true]{standalone}
\usepackage{nicefrac}
\usepackage{xcolor}

\usepackage{mystyle}

\AtBeginDocument{%
  }

\setcopyright{acmlicensed}
\copyrightyear{2024}
\acmYear{2024}
\acmDOI{XXXXXXX.XXXXXXX}

\acmJournal{POMACS}
\acmVolume{37}
\acmNumber{4}
\acmArticle{111}
\acmMonth{8}

\acmSubmissionID{123-A56-BU3}

\begin{document}

\title[Imitation Learning Survey]{A Survey of Imitation Learning Methods, Environments and Metrics}

\author{Nathan Gavenski}
\email{nathan.schneider_gavenski@kcl.ac.uk}
\orcid{0000-0003-0578-3086}
\affiliation{
    \institution{King's College London}
    \city{London}
    \country{United Kingdom}
}

\author{Felipe Meneguzzi}
\orcid{0000-0003-3549-6168}
\affiliation{%
  \institution{University of Aberdeen}
  \city{Aberdeen}
  \country{United Kingdom}
}

\author{Michael Luck}
\orcid{0000-0002-0926-2061}
\affiliation{%
  \institution{University of Sussex}
  \city{Sussex}
  \country{United Kingdom}
}

\author{Odinaldo Rodrigues}
\orcid{0000-0001-7823-1034}
\affiliation{%
  \institution{King's College London}
  \city{London}
  \country{United Kingdom}
}

\renewcommand{\shortauthors}{Gavenski et al.}

\begin{abstract}
    Imitation learning is an approach in which an agent learns how to execute a task by trying to mimic how one or more \textit{teachers} perform it. 
    This learning approach offers a compromise between the time it takes to learn a new task alone by trial-and-error and the effort needed to collect the large number of samples needed for supervised learning.
    The idea is to focus on the samples provided by the teacher, who is assumed to be proficient at executing the task, deviating from the samples only when faced with unseen situations. 
    This is especially useful when we want to imbue an agent with a desired behaviour that is difficult to
    specify through a particular reward function.
    As a result, the field of imitation learning has received considerable attention from researchers, resulting in many new techniques and applications. 
    This sudden increase in publications poses some challenges when it comes to comparing work and evaluating results in a uniform manner. 
    In this survey, we tackle these challenges by systematically reviewing the imitation learning literature and presenting our findings by 
    \begin{enumerate*}[label=(\roman*)]
        \item introducing novel taxonomies which focus on critical aspects of the imitation learning methods, environments and metrics;
        \item reflecting and discussing important issues observed across publications; and
        \item highlighting open challenges and suggesting directions for future work.
    \end{enumerate*}
    This survey provides a solid foundation for understanding the increasingly diverse and dynamic imitation learning field.
    It allows the development of new approaches and a more systematic and uniform comparison of all techniques.
\end{abstract}




\maketitle

\section{Introduction} \label{sec:introduction}

Imitation learning is a socially-inspired machine learning approach that consists of learning a task from examples provided by a teacher.
In other words, this approach refers to an agent's acquisition of skills or behaviours by observing a teacher performing a given task~\cite{hussein2017imitation}.
This approach benefits the agent since it learns from samples that theoretically indicate a successful behaviour instead of searching which behaviours solve the task (a process that trial-and-error learning techniques have to go through).
Moreover, this idea from imitation learning is compelling since, when learning something new, the learner tries to learn from a proficient source and adopt more expert-like behaviour, which allows for more human-centric approaches.

In recent years, imitation learning evolved significantly from \citeauthor{pomerleau1988alvinn}'s original work on behavioural cloning~\cite{pomerleau1988alvinn}.
Some work expanded the application of imitation learning with more complex learner structures, such as ensemble methods~\cite{kidambi2021mobile}.
Other work proposed novel learning schematics, such as adversarial learning~\cite{ho2016generative} and self-supervised learning~\cite{torabi2018bco}.
These approaches rely on machine learning techniques to optimise an agent's behaviour, while other researchers focused on agent techniques by proposing the usage of exploration~\cite{gavenski2020iupe} and remapping of latent actions~\cite{edwards2019ilpo}.
Given these improvements, imitation learning has seen many applications, including robotics~\cite{zhang2018deep,pathak2018zeroshot,nair2017combining,lynch2020learning,finn2016guided,yang2023watch,paolillo2023dynamical,rodrigo2023stable,wen2019efficient,liu2018raw}, game-playing~\cite{shih2022conditional,shang2021self,monteiro2020abco,gavenski2020iupe,zhu2020opolo,yu2020giril,edwards2019ilpo,guo2019hybrid,peng2018variational,yusuf2018playing,pathak2018zeroshot,lee2014mario,ortega2013mario,ross2011reduction}, and natural language processing~\cite{zhao2022cacti,pulver2021pilot,li2019divine,du2019empirical,vlachos2017imitation}.

With this increase in popularity, imitation learning has seen a rise in the number of experimental environments and performance metrics, as we show in this survey.
These new metrics and environments, along with other surveys~\cite{hussein2017imitation,torabi2019advances,zheng2021imitation} focusing more on learning techniques, create a lack of standardisation in the evaluation process, making meaningful comparisons between different approaches challenging.
Hence, this survey examines the field from a different perspective from the traditional model-free and model-based views employed in the past.
It presents the field based on the most predominant learning techniques, metrics and environments, including the first taxonomies for environments and metrics and a new methodological one.
We believe that our taxonomy for methods does not nullify existing methodological taxonomies but rather complements them by classifying novel trends in imitation learning methods in more detail.
This paper thus acknowledges prior contributions from previous surveys and incorporates them when appropriate or necessary.

We first describe the selection criteria for including the various research efforts in the systematic review and highlight the main differences with previous reviews in Section~\ref{sec:review_preliminaries}. 
Section~\ref{sec:background} lays out the formal underpinning of imitation learning, defining the critical elements involved in imitation learning tasks, and allowing our discussion to remain mathematically rigorous. 
We then present the different approaches to imitation learning in Section~\ref{sec:methods}, introducing a new taxonomy to highlight a new trend among learning approaches. 
In Section~\ref{sec:environments}, we present the different environments used in imitation learning and propose the first taxonomy for imitation learning environments based on the {\em role} of the environment in the learning process.
Section~\ref{sec:metrics} explains the different metrics used to evaluate imitation learning approaches, groups them based on each of their measures, and presents the first taxonomy for imitation learning metrics. 
In Section~\ref{sec:reflections}, we discuss key insights derived from the literature, pose key challenges for the field, and possible future directions for imitation learning.
Finally, Section~\ref{sec:conclusion} summarises our main contributions by emphasising the need for consistent use of evaluation processes and environment and how researchers can use these new taxonomies to achieve a more solid and systematic~approach.
\section{Review preliminaries} \label{sec:review_preliminaries}

In this work, we use a snowballing approach~\cite{snowballing2014Wohlin} and two different surveys: 
\citeauthor{hussein2017imitation}~\cite{hussein2017imitation} from $2017$ and \citeauthor{zheng2021imitation}~\cite{zheng2021imitation} from $2021$ as a starting point.

We select \citeauthor{hussein2017imitation}'s work since it cites significant work on imitation learning and defines various approaches used to learn from demonstrations.
With more than $800$ citations, the snowballing process is bound to find the most relevant work to our desired subject.
Given that the first survey is older, we add \citeauthor{zheng2021imitation} as a more recent survey.
By adding a more novel survey, we are more assured of finding relevant work that we might miss from an older survey in a snowballing process.
To select all relevant work, we follow two general guidelines when filtering all papers found during the snowballing process.
The first criterion is that all work must be from the inverse reinforcement learning or imitation learning literature because, historically, imitation learning approaches appeared in the reinforcement learning literature.
Furthermore, imitation learning approaches that use inverse reinforcement learning require teacher demonstrations to be considered imitation learning, we do not exclude any inverse reinforcement learning method without further inspection.
The second inclusion criterion refers to the publishing venues.
We only use published peer-reviewed work, except for pre-prints published since $2023$, given the slow nature of the rigorous review process for machine learning publication.
This exception provides just one addition (\citeauthor{rodrigo2023stable}~\cite{rodrigo2023stable}).
Although these rules set a general framework for selecting relevant work for our literature review, they are just guidelines that should not be seen as strict rules.

We use abstracts to remove all work that uses imitation learning outside the context of agent-related research.
Examples of this are papers on meta-learning, which use imitation learning techniques applied to other domains, such as neural networks training~\cite{zhen2023repairing}.
From the later survey~\cite{zheng2021imitation}, we present $30$ new methods, which shows the fast-growing nature of the imitation learning field.
For completeness, we include other surveys (which others~\cite{torabi2019advances} omit) since imitation learning takes inspiration from other fields, such as reinforcement learning~\cite{arulkumaran2017reinforcement}, multi-agent learning~\cite{du2020multiagent,busoniu2008multiagent} and transfer learning~\cite{zhuang2021transfer,pan2010transfer}.

To help bring some uniformity to our review, we deviate slightly in approach from previous surveys and create a unifying taxonomy under which different categories of environments and metrics can be understood and compared. 
Hence, we consider the field from a different perspective to the traditional model-free and model-based view employed by \citeauthor{torabi2019advances}~\cite{torabi2019advances}.
We attribute the lack of standardisation in the imitation learning field to the lack of a taxonomy for imitation learning environments and metrics, which is the main reason for us to deviate from past surveys.
In \citeauthor{hussein2017imitation}~\cite{hussein2017imitation}, the authors mainly focused on methodologies, given the need for classifying new approaches from the literature.
They set a background for future work, from which we take inspiration, and introduce relevant research following their learning approach, such as apprenticeship learning, behavioural cloning, and inverse reinforcement learning.
We follow a similar approach for our classification of imitation learning methods, in which we present a taxonomy based on the most predominant learning techniques during training, such as inverse dynamics models and adversarial learning.
Given its simplistic nature for classifying imitation learning methods, we deviate from \citeauthor{zheng2021imitation}, who classify each approach as either a behavioural cloning, inverse reinforcement learning or adversarial imitation learning approach or as high-level (which requires higher-level cognitive functions) or low-level (concrete operations, such as moving an object) tasks.
Moreover, we maintain that our classification system not only acknowledges their taxonomy but also enhances it by providing a more detailed categorisation of emerging trends within imitation learning methods.

Finally, we present a novel taxonomy for imitation learning environments based on the role of the environment in the learning process and a taxonomy for imitation learning metrics based on the meaning they convey.
Although \citeauthor{hussein2017imitation} and \citeauthor{zheng2021imitation} discuss different imitation learning environments, they do not classify them and only group them by domain, which provides no insight into the environment's role in the learning process.
Similarly to environments, \citeauthor{hussein2017imitation} present different metrics for evaluating imitation learning methods. 
However, their work does not classify them nor present in-depth metrics analysis.
\section{Background} \label{sec:background}

Imitation learning refers to an \textit{agent}'s acquisition of skills or behaviours by observing a source performing a given task~\cite{hussein2017imitation}.
In the imitation learning context, we follow \citeauthor{norvig2002modern}'s definition~\cite[Chapter 3.]{norvig2002modern} of an agent where an \textit{agent} is an entity that autonomously interacts within an environment to achieve a goal.
We refer to these sources of behaviour as \textit{teachers}.
Unlike other agent-based learning methods, where an agent learns through interactions with the environment, imitation learning uses this teacher's behaviours to guide the learning process.
Most commonly, imitation learning refers to the teacher as an `expert', but we deviate from this nomenclature, which we further discuss the reasoning behind in Section~\ref{sec:reflections}.
In this setting, a teacher can be a human or a computational agent.
We generally define a teacher as an agent who provides information on how to act in an environment.
This information might carry expert knowledge, but it is not a requirement.

More formally, imitation learning problems use \textit{Markov Decision Processes} to model the environment.
This formulation represents a state-action network, where the transition of states is mapped from its state-actions and, therefore, is suitable for imitation learning.
A more formal definition of the Markov decision process is:
\begin{definition} \label{def:mdp}
    \textit{Markov Decision Processes} are represented by a five-tuple $\mdp = \langle \MDPstate, \action, T, r, \gamma \rangle$, in which: $\MDPstate$ is the state space, $\action$ is the action space, $T$ is the transition dynamics $T : \MDPstate \times \action \rightarrow \MDPstate$, $r$ is the immediate reward function $r : \MDPstate \times \action \rightarrow \mathbb{R}$, and $\gamma$ is the discount factor such that $\gamma \in [0,1)$.
\end{definition}

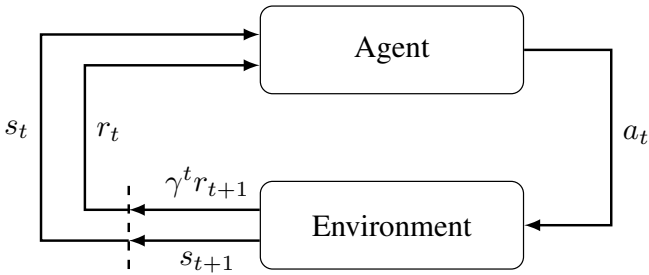
\begin{figure}[htb]
    \centering
    \begin{tikzpicture}[
    node/.style={
        rectangle, 
        rounded corners, 
        minimum width=3cm, 
        minimum height=1cm,
        text centered, 
        draw=black
    },     
    arrow/.style={
        thick,
        ->,
        >=latex
    }
]
    \node at (\agentpos, 2.5) (agent) [node] {Agent};
    \node at (\env, 2.5) (env) [node] {Environment};

    \draw [thick, dashed] (4.25, 2) -- (4.25, 3);

    \draw [arrow] 
        (agent.east)  
        -| (\agentpos+2, 3.4)
        -- (\env-2, 3.4) node[pos=.5,above] {$a_{t}$}
        |- (env.west);
    
    \draw [arrow, -] 
        ([yshift=5pt]env.east) 
        -- ([yshift=5pt]4.2, 2.5) 
        node[pos=0.5,above] {$\gamma^t r_{t+1}$};
    \draw [arrow] 
        ([yshift=5pt]4.3, 2.5) 
        -- ([yshift=5pt]agent.west)
        node[pos=.5,above] {$r_{t}$};
    
    \draw [arrow, -] 
        ([yshift=-5pt]env.east) 
        -- ([yshift=-5pt]4.2, 2.5) 
        node[pos=.5,below] {$s_{t+1}$};
    \draw [arrow] 
        ([yshift=-5pt]4.3, 2.5) 
        -- ([yshift=-5pt]agent.west)
        node[pos=.5,below] {$s_{t}$};
\end{tikzpicture}
    \caption{The Markovian Decision Process.}
    \label{fig:mdp_diagram}
    \Description{Markovian Decision Process diagram.}
\end{figure}

\noindent
The state space $\MDPstate$ is represented by vectors that contain information regarding the environment's current state.
For the action space $\action$, we can also encounter two forms to represent them: vector-based and single values.
Vector-based actions represent either a sequence of decisions that the agent should enact in order, such as a sequence of buttons to press, or a list of actions to act simultaneously, different joints from a robot, for example.
The transition dynamics of a Markov Decision Process dictate how a transition in an environment will occur, while the discount factor relates to the reward regarding the time domain.
A discount factor of $1$ indicates that future rewards are considered equally important as immediate rewards, while a discount factor close to $0$ implies that only immediate rewards are prioritised.
Moreover, a critical property of these networks is that the transition dynamics usually only depend on the previous state and action, regardless of earlier states.
Nevertheless, in the context of imitation learning, although the network might carry information regarding reward and discount factors, we consider that this information is inaccessible to the agent, and the learning process does not depend on it.
Fig.~\ref{fig:mdp_diagram} displays a common implementation of the Markovian problem where the environment provides a state $s$ to the agent, which predicts the most likely action $a$.
The environment applies the state transition function $T$ on the state-action pair to generate the new state $s'$, and computes the immediate reward using the discount factor $\gamma$ and immediate reward function $r$.

Following a Markov Decision Process definition, we can generally state that the goal of imitation learning is for an agent to use $\MDPstate$ and $\action$ from a teacher to learn a mapping function called \textit{policy}.
A teacher also has a known or unknown policy that the agent is trying to approximate, and we define this function as:

\begin{definition}
    A \textit{policy} is a function $\policy$ that maps states to actions $\policy := \MDPstate \rightarrow \action$.
    The policy has inner parameters $\theta$, which represent the internal variables or weights within the policy that are adjusted during the learning process. 
    The set of all possible internal variables is defined as $\Pi = \left \{ \pi_1, \pi_2, \dots, \pi_z \right \}$.
\end{definition}

\noindent
Strictly defined, a policy maps states to probabilities of selecting each possible action $P(a \mid s)$ conditioned by the available teacher's data.
Having the MDP definition allows us to define agents and teachers more formally.

\begin{definition}
    An \textit{agent} $\agent$ is one possible instantiation of $\Policy$.
    It selects an action $a \in \action$ provided an environment state $s \in \MDPstate$ according to its \textit{fixed learned} parameters $\theta$.
\end{definition}

\begin{definition}
    A \textit{teacher} $\teacher$ is a special case of an agent, with respect to its parameters $\psi$, whose behaviour imitation learning wants to approximate.
\end{definition}

\noindent
Applying an action $a$ to the environment results in a new state representation $s'$, which an agent can use to select a new action, forming a sequence of states and actions.
All possible sequences from a given \textit{teacher} refer to a smaller of all possible environment states $\MDPstate$ and actions $\action$, which we denote as $S_\teacher$ and $A_\teacher$, respectively.
These states and actions are subsets because solutions from $\teacher$ may not require it to visit a specific state or predict an action.
Thus, $S_\teacher \subseteq \MDPstate$ and $A_\teacher \subseteq \action$.
For simplicity, in this work, we omit the policy information when referring to $\teacher$, and explicitly inform the policy otherwise.

We still lack a definition for the sequential data collected from teachers to learn the policy function.
These sequences of state and action interactions from the teacher in the environment are called \textit{trajectories} and are an essential part of the imitation learning processes.
The data within each trajectory is the only information an imitation learning agent can access when learning how to perform a task in an environment.
Therefore, if too few samples are collected, the agent might encounter states with no prior information on how to act, or if the data is corrupted, it may act differently than its teacher.
In this report, we formally define trajectories as:

\begin{definition}
    A \textit{trajectory} is an ordered list of state-action pairs $\tau = \left [ (s_1, a_1), (s_2, a_2), \dots, (s_n, a_n) \right ]$, where $n > 1$ and $s_{i+1} = T(s_i, a_i)$, for all $i > 1$.
\end{definition}

\noindent
A set of trajectories is composed of multiple trajectories, such that $\Tau = \left \langle \tau_1, \tau_2, \dots, \tau_k \right \rangle$, where $k \geqslant 1$.
Given a set of trajectories $\Tau$, an agent $\agent$ and a teacher $\teacher$, we can mathematically express the learning process for imitation learning as minimising the error for a given loss function $\loss$ between $\teacher$ and $\agent$, as follows:

\begin{equation} \label{eq:il_goal}
    \argmin_\theta \sum_{\tau \in \Tau} \sum_{s \in \tau} \loss(\teacher(s), \agent(s)),
\end{equation}

\noindent
where $s$ is a state sampled from the teacher's trajectory $\tau$ in the set of trajectories $\Tau$.
It is important to note that early iterations of imitation learning used supervised learning losses to learn a policy. 
However, novel work~\cite{gavenski2020iupe,kidambi2021mobile} have shown that imitation learning can be formulated as self-supervised and adversarial learning approaches (and further discussed in Section~\ref{sec:methods}).

Equation~\ref{eq:il_goal} assumes we have access to $\teacher$.
However, it is sensible to assume that imitation learning could not depend on direct access to a teacher's policy.
This level of access requires knowledge of the teacher's internal state, which cannot be done when the teacher is human.
Therefore, imitation learning may use \textit{demonstrations}, also known as \textit{imitation learning from demonstrations}, to learn a policy, which we define below.

\begin{definition}
    A \textit{demonstration} is a state-action pair taken from a teacher's trajectory, such that $(s, a) \sim \tau \in \Tau$.
\end{definition}

\noindent 
A list of all sampled demonstrations from one or more teachers is denoted as $D = \left [ (s_1, a_1), \dots, (s_n, a_n) \right ]$.
We denote the subset of all states for a demonstration is $S = \left \{ s \mid (s, a) \in D \text{ for some } a \right \}$, while the subset of all actions is $A = \left \{ a \mid (s, a) \in D \text{ for some } s \right \}$.
The probability of taking an action $a$ given $s$ in a list of demonstration $D$ is:

\begin{figure}[h]
    \centering
    \begin{minipage}{.49\linewidth}
        \begin{equation}\label{eq:prob_D}
            P_D(s, a) = 
            \left\{\begin{matrix}
                P(s, a) & \text{if } (s, a) \in D \\ 
                0 & \text{otherwise}
            \end{matrix}\right., \text{ where }
        \end{equation}
    \end{minipage}
    \hfill
    \begin{minipage}{.49\linewidth}
        \begin{equation}\label{eq:prob_d}
            P(s, a) = \frac{
                \left| \left \{ (s_i, a_i) \in D \mid s_i = s, a_i = a \right\} \right|
            }{
                \left| \left \{ (s_i, a) \in D \mid s_i = s, \text{for some~}a \right\} \right|
            }.
        \end{equation}
    \end{minipage}
\end{figure}

\noindent
Unlike trajectories, demonstration lists are not necessarily ordered lists; hence, $s_{i+1}$ may not be the direct outcome from $T(s_i, a_i)$.
As trajectories, demonstrations are vital for imitation learning and highly correlate with the agent's ability to generalise in unseen scenarios.
For example, when solving different puzzles, the agent must adapt its strategy to solve one puzzle where some pieces do not share the same shapes from previous demonstrations.
Consequently, when creating imitation learning from demonstrations, researchers must consider whether the data is representative enough of the environment, which we further discuss in Section~\ref{sec:reflections}.

Although imitation learning shies away from any direct signal from an environment by using demonstration, \textit{experiences} such as an agent's trajectories can also be used. 
The difference between the two instances of data is that demonstrations provide the teacher's action to a given state, allowing a supervised learning approach, and experiences show the performed action, which may not be close to the source behaviour but also provides the reward of performing that action given the current state.
However, using the reward function to optimise its policy further goes against the imitation learning paradigm.
Hence, these work avoid using the returned reward from the environment and use experiences as state-action pairs.
More formally, we define experiences as follows.

\begin{definition} \label{def:exp}
    An \textit{experience} is a special case of a demonstration, in the sense that its tuples are from the learned agent's trajectories: $(s, a) \sim \tau \in \Tau_\agent$ instead of the teacher's.
\end{definition}

All definitions so far help us understand early approaches to imitation learning.
Conversely, these definitions do not align with how humans acquire knowledge through observation.
Thus, \textit{imitation learning from observation} further restricts imitation learning by reducing demonstration information only to the teacher's states.
By learning without knowing which actions were performed, imitation learning from observation tries to employ the same learning approach humans do by figuring out the action performed from the observed changes between two successive states of the environment.
In this context, imitation learning from observation represents its examples with less information.

\begin{definition}
    An \textit{observation} is a state pair $(s_i, s_{i+1})$ from a teacher.
\end{definition}

\begin{figure}[htb]
    \centering
    \tikzstyle{node} = [
    rectangle,
    thick,
    rounded corners, 
    minimum width=3cm, 
    minimum height=7mm,
    text centered, 
    draw=black
]
\tikzstyle{arrow} = [->,>=latex]
\begin{tikzpicture}[
    node distance=2cm,
    node/.style={
        rectangle, 
        rounded corners, 
        minimum width=3cm, 
        minimum height=1cm,
        text centered, 
        draw=black
    },     
    arrow/.style={
        ->,
        >=latex
    }
]
    
    \node [align=left] at (2, 8.5) {Training for IL agents};

    \node at (0.8958333333, 7.125) (state) {$s_i$};

    \node (teacher) [node, minimum height=.75cm, minimum width=2cm] at (4.479166667, 6.4375) {$\text{Teacher }{\teacher}$};
    \node (agent) [node, minimum height=.75cm, minimum width=2cm] at (4.479166667, 7.8125) {$\text{Agent }{\agent}$};
    \node (loss) [node, minimum height=.75cm, minimum width=2cm] at (8.0625, 7.125) {$\text{loss }{\loss}$};

    \draw [arrow] (state) -- (1.791666667, 7.125) |- (teacher.west);
    \draw [arrow] (state) -- (1.791666667, 7.125) |- (agent.west);
    \draw [arrow] (agent.east) -- (6.270833333, 7.8125) |- node[pos=.1, yshift=10pt, xshift=0pt] {$a_\agent$} (loss);
    \draw [arrow] (teacher.east) -- (6.270833333, 6.4375) |- node[pos=.1, yshift=-11pt, xshift=0pt] {$a_\teacher$} (loss);

    \draw [arrow] (loss.east) -- (9.854166667, 7.125) |- (9.854166667, 8.5) -| node[yshift=7pt, xshift=80pt] {$ \text{update } \theta$} (agent.north);

    \draw [thick, dash dot] (10.75, 5.75) -- (10.75, 9);

    \node [align=left] at (14.1, 8.5) {Evaluation for IL agents};    
    \node at (14, 7.125) (env) [node] {$\text{Environment}$};
    \node at (20, 7.125) (agent) [node] {$\text{Agent }{\agent}$};
    \draw [thick, dashed] (17, 6.5) -- (17, 7.75);
    
    \draw [arrow] ([yshift=-0.2cm]env.east) --
                  node[pos=.25, anchor=north] {$s_{t+1}$}
                  node[pos=.75, anchor=north] {$s_{t}$}
                  ([yshift=-0.2cm]agent.west);
    \draw [arrow] ([yshift=0.2cm]env.east) --
                  node[pos=.25, anchor=south] {$\gamma^{t\;} r_{t+1}$}
                  node[pos=.75, anchor=south] {$r_{t}$}
                  ([yshift=0.2cm]agent.west);
    \draw [arrow] (agent.east) -|
                  ++(0.5, -1) --
                  node[anchor=north] {$\text{action } a_t$}
                  ++(-10, 0) --
                  ++(0, 1) --
                  (env.west);



\end{tikzpicture}
    \caption{A general approach to imitation learning.}
    \label{fig:il_diagram}
    \Description{Diagram for imitation learning approaches.}
\end{figure}
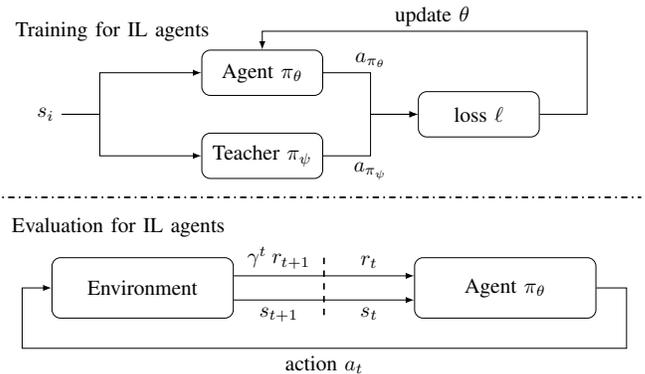

\noindent
If we have access to the teacher's trajectory $\tau$ and the transition function $T$, we can retrieve any observation with $(s_i, T(s_i, a_i))$, such that $(s_i, a_i) \in \tau$ and $i \leqslant n$.
On the other hand, if we lack these pieces of information, we can only retrieve observations $(s_i, s_{i+1})$ for all sequences of states where $i < n$.
A list of observations is a not necessarily ordered list of all sampled observations from $\teacher$: $O = \left [ (s_1, s_2), \dots, (s_{n-1}, s_n) \right ]$, where $n > 1$.
Like demonstrations, the subset of states for observations is as follows: $\left \{ s \mid (s_i, s_{i+1}) \in O \right \}$.
The probability for observation pairs follows Equations~\ref{eq:prob_D} and~\ref{eq:prob_d}, but uses the observation's information rather than the demonstration's.
Assuming access only to the state information helps because datasets that explicitly give the actions performed between state changes (also known as \textit{labelled datasets}) are uncommon in the real-world\footnote{This is commonly referred to as \textit{in the wild}} and are costly to create.
By lifting this restriction, imitation learning agents do not need to create a dataset that involves recording humans playing or training another agent to act as a teacher.
However, by removing the action information, two problems may arise when dealing with more complex tasks:
\begin{enumerate*}[label=(\roman*)]
    \item Markovian problems often are non-injective, which means that the transition function $T$ may map two different state-action pairs to the same state, and smaller datasets will not cover the range of transitions necessary for the policy to learn properly; or
    \item computing whether the collected data covers a significant part of the Markovian network might be impossible.
\end{enumerate*}
Therefore, focusing on imitation learning from observation is vital to create more efficient imitation learning agents and removing the requirement for vast datasets.
Fig.~\ref{fig:il_diagram} shows a general approach to imitation learning, where the agent learns from demonstrations or observations provided by a teacher and uses the environment to collect new experiences.

So far, we have described all possible inputs for imitation learning agents as time-sensitive, but our agent formulation for now remains time-independent.
Thus, we have to consider time as a possible input for all types of data, despite its ability to specify an instance of input and output.
A policy that maps a sequence of states to actions, such as $P(A \mid s_0, s_1, \cdots, s_t)$, is called \textit{non-stationary}. 
These policies help map lasting consequences between states and sequential actions, such as the agent closing a path unintentionally and the different movements required to hit a ball with a racquet, respectively.
Therefore, non-stationary policies are more naturally suited to learning motor trajectories~\cite{ross2011reduction}.
Conversely, \textit{stationary} policies ignore time and predict actions solely on the present information $P(A \mid s)$.
The advantage of stationary policies is the ability to learn tasks when their length might be too long or unknown~\cite{ross2011reduction}, such as autonomous driving, where the agent can drive for an undetermined time.
Moreover, non-stationary policies are difficult to adapt to unseen scenarios and changes in the parameters of a task~\cite{schaal2003computational}.
Given the temporal characteristics of these policies, at one point, the trajectory can result in compounded errors as the agent continues to perform the remainder of the actions, such as an agent moving in a loop through a maze.
The imitation learning literature focuses heavily on stationary agents.

Lastly, \citeauthor{zheng2021imitation}~\cite{zheng2021imitation} propose to further classify actions into two groups:
\begin{enumerate*}[label=(\roman*)]
    \item \textit{low-level} actions, which refer to atomic decisions that an agent can perform in a given domain, typically involving simple commands such as movement or interaction; or
    \item \textit{high-level} actions that are decisions made by an agent that determines the overall plan or strategy to be executed in a given task or domain, which are often complex and involve multiple lower-level actions or sequences of decisions. 
\end{enumerate*}
Most imitation learning work focuses on low-level actions of single-value outputs, given the stationary nature of most agents.

\begin{figure*}[t]
    \centering
    \begin{tikzpicture}[
    node distance=3.5cm,
    every node/.style = {
        rectangle, 
        rounded corners,
        minimum height=1cm,
        font=\large \bfseries,
        text centered
    },
    node/.style = {
        minimum width = 5cm
    },
    subnode/.style = {
        minimum width = 4cm
    },
    arrow/.style = {
        ultra thick,
        ->,
        >=latex
    }
]
    \path[use as bounding box] (-15.65,-7.1) -- (19.35,0.6);
    \node (imitation_learning) [
        node, fill=white!10, draw=black, minimum width={width("Imitation Learning Methods (50)")+100pt}
    ] {Imitation Learning Methods (50)};

    \node (adversarial) [node, below of=imitation_learning, fill=adversarial!15, draw=adversarial] {Adversarial Learning (15)};
    \node (inverse_reinforcement) [%
        subnode, 
        below of=adversarial, 
        fill=adversarial!15, 
        draw=adversarial, 
        node distance=1.5cm, 
        xshift=1.4cm,
        align=center
    ] {Inverse Reinforcement \\Learning (4)};
    \node (adversarial_imitation_learning) [%
        subnode, 
        below=of inverse_reinforcement.west, 
        fill=adversarial!15, 
        draw=adversarial, 
        yshift=2cm,
        anchor=west,
        align=center
    ] {Adversarial Imitation \\Learning (11)};

    \node (dynamics) [node, left of=adversarial, xshift=-3cm, fill=dynamics!10, draw=dynamics!50] {Dynamics Model methods (14)};
    \node (idm) [%
        subnode, 
        below of=dynamics, 
        fill=dynamics!10, 
        draw=dynamics!50, 
        node distance=1.5cm, 
        xshift=1cm,
        align=center
    ] {Inverse Dynamics \\Models (10)};
    \node (fdm) [%
        subnode, 
        below=of idm.west,
        fill=dynamics!10,
        draw=dynamics!50,
        yshift=2cm,
        anchor=west,
        align=center
    ] {Forward Dynamics \\Models (4)};

    \node (hybrid) [node, right of=adversarial, xshift=3cm, fill=hybrid!25, draw=hybrid] {Hybrid methods (8)} ;
    
    \node (behaviour) [node, left of=dynamics, xshift=-3cm, fill=bc!25, draw=bc] {Behaviour Cloning (11)};

    \node (online) [node, right of=hybrid, xshift=3cm, fill=violet!15, draw=violet] {Online Learning (2)};
    \node (ftl) [%
        subnode,
        below of=online,
        fill=violet!15,
        draw=violet,
        node distance=1.5cm,
        xshift=1.3cm
    ] {Follow-the-Leader (2)};
    \node (ftrl) [%
        subnode,
        below=of ftl.west,
        fill=violet!15,
        draw=violet,
        yshift=2cm,
        anchor=west
    ] {Follow-the-Regularized-Leader (1)};

    \draw [arrow]   (imitation_learning) -- (adversarial);
    \draw [arrow]   ++(0, -1.25cm) -|
                    (dynamics.north);
    \draw [arrow]   ++(0, -1.25cm) -|
                    (behaviour.north);
    \draw [arrow]   ++(0, -1.25cm) -|
                    (hybrid.north);
    \draw [arrow]   ++(0, -1.25cm) -|
                    (online.north);

    \draw [arrow]   ([xshift=-1.5cm]adversarial.south) |- (inverse_reinforcement.west);
    \draw [arrow]   ([xshift=-1.5cm]adversarial.south) |- (adversarial_imitation_learning.west);
    \draw [arrow]   ([xshift=-1.5cm]dynamics.south) |- (idm.west);
    \draw [arrow]   ([xshift=-1.5cm]dynamics.south) |- (fdm.west);
    \draw [arrow]   ([xshift=-1.5cm]online.south) |- (ftl.west);
    \draw [arrow]   ([xshift=-1.5cm]online.south) |- (ftrl.west);

\end{tikzpicture}
    \caption{A new taxonomy for imitation learning methods.}
    \label{fig:taxonomy}
    \Description{Proposed taxonomy for imitation learning methods.}
\end{figure*}
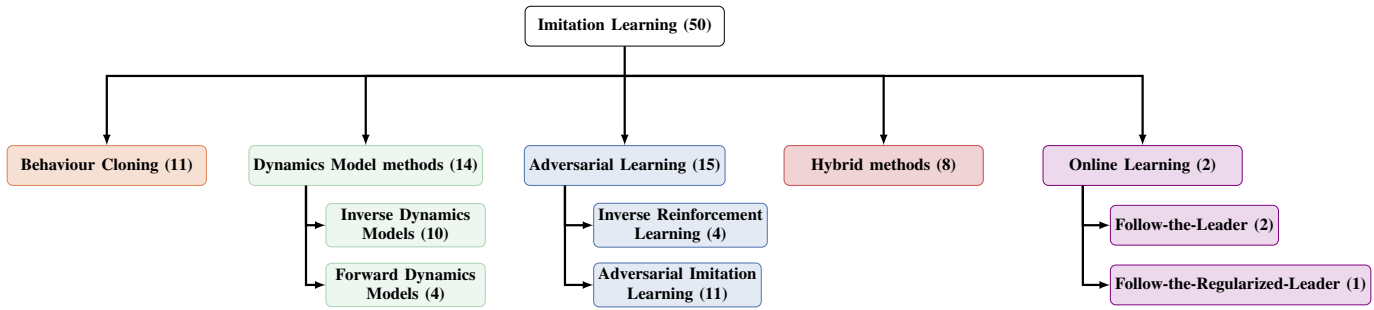

\section{Imitation Learning Methods} \label{sec:methods}

Common forms of classification for imitation learning partition the methods into model-based and model-free methods.
Model-based methods rely on learning to model the environment, such as predicting the consequences of
\begin{wrapfigure}{l}{0.35\linewidth}
    \centering
    \begin{tikzpicture}
    \draw[color=black, thick] (0.25, 0.25) -- (8, 0.25); 
    \draw[color=black, thick] (0.25, 0.25) -- (0.25, 9.5); 

    \node[below=0.2cm] at (2, 0.25) {Model-free};
    \node[below=0.2cm] at (6, 0.25) {Model-based};

    \node[below=0.5cm, font=\bfseries] at (4.25, 0) {Traditional Taxonomy};

    \node[circle, fill=adversarial, minimum size=\myminsize] at (1,1) {\textcolor{white}{\citenum{chang2022ilflow}}};
    \node[circle, fill=adversarial, minimum size=\myminsize] at (2,1) {\textcolor{white}{\citenum{torabi2018gaifo}}};
    \node[circle, fill=adversarial, minimum size=\myminsize] at (3,1) {\textcolor{white}{\citenum{yin2022planning}}};
    \node[circle, fill=adversarial, minimum size=\myminsize] at (1,2) {\textcolor{white}{\citenum{shang2021self}}};
    \node[circle, fill=adversarial, minimum size=\myminsize] at (2,2) {\textcolor{white}{\citenum{wen2019efficient}}};
    \node[circle, fill=adversarial, minimum size=\myminsize] at (3,2) {\textcolor{white}{\citenum{peng2018variational}}};
    \node[circle, fill=adversarial, minimum size=\myminsize] at (1,3) {\textcolor{white}{\citenum{hu2023safe}}};
    \node[circle, fill=adversarial, minimum size=\myminsize] at (2,3) {\textcolor{white}{\citenum{finn2016guided}}};
    \node[circle, fill=adversarial, minimum size=\myminsize] at (3,3) {\textcolor{white}{\citenum{yunzhu2017infogail}}};
    \node[circle, fill=adversarial, minimum size=\myminsize] at (1,4) {\textcolor{white}{\citenum{jiaming2018multiagent}}};
    \node[circle, fill=adversarial, minimum size=\myminsize] at (2,4) {\textcolor{white}{\citenum{torabi2019video}}};
    \node[circle, fill=adversarial, minimum size=\myminsize] at (3,4) {\textcolor{white}{\citenum{lee2014mario}}};
    \node[circle, fill=adversarial, minimum size=\myminsize] at (1,5) {\textcolor{white}{\citenum{shang2021self}}};
    \node[circle, fill=adversarial, minimum size=\myminsize] at (2,5) {\textcolor{white}{\citenum{shih2022conditional}}};
    \node[circle, fill=adversarial, minimum size=\myminsize] at (3,5) {\textcolor{white}{\citenum{monteiro2023self}}};

    \node[circle, fill=bc, minimum size=\myminsize] at (1,6) {\textcolor{white}{\citenum{chen2020cheating}}};
    \node[circle, fill=bc, minimum size=\myminsize] at (2,6) {\textcolor{white}{\citenum{yan2017oneshot}}};
    \node[circle, fill=bc, minimum size=\myminsize] at (3,6) {\textcolor{white}{\citenum{lynch2020learning}}};
    \node[circle, fill=bc, minimum size=\myminsize] at (1,7) {\textcolor{white}{\citenum{zhang2018deep}}};
    \node[circle, fill=bc, minimum size=\myminsize] at (2,7) {\textcolor{white}{\citenum{raza2012teaching}}};
    \node[circle, fill=bc, minimum size=\myminsize] at (3,7) {\textcolor{white}{\citenum{dambrosio2013scalable}}};
    \node[circle, fill=bc, minimum size=\myminsize] at (1,8) {\textcolor{white}{\citenum{chernova2008teaching}}};
    \node[circle, fill=bc, minimum size=\myminsize] at (2,8) {\textcolor{white}{\citenum{calinon2007what}}};
    \node[circle, fill=bc, minimum size=\myminsize] at (3,8) {\textcolor{white}{\citenum{michael1999bc}}}; 
    \node[circle, fill=bc, minimum size=\myminsize] at (1,9) {\textcolor{white}{\citenum{daftry2017mav}}};
    \node[circle, fill=bc, minimum size=\myminsize] at (2,9) {\textcolor{white}{\citenum{carroll2019overcooked}}};

    \node[circle, fill=dynamics, minimum size=\myminsize] at (5,1) {\textcolor{white}{\citenum{torabi2018bco}}};
    \node[circle, fill=dynamics, minimum size=\myminsize] at (6,1) {\textcolor{white}{\citenum{edwards2019ilpo}}};
    \node[circle, fill=dynamics, minimum size=\myminsize] at (7,1) {\textcolor{white}{\citenum{monteiro2020abco}}};
    \node[circle, fill=dynamics, minimum size=\myminsize] at (5,2) {\textcolor{white}{\citenum{gavenski2020iupe}}};
    \node[circle, fill=dynamics, minimum size=\myminsize] at (6,2) {\textcolor{white}{\citenum{gavenski2021self}}};
    \node[circle, fill=dynamics, minimum size=\myminsize] at (7,2) {\textcolor{white}{\citenum{raychaudhuri2021cross}}};
    \node[circle, fill=dynamics, minimum size=\myminsize] at (5,3) {\textcolor{white}{\citenum{rodrigo2023stable}}};
    \node[circle, fill=dynamics, minimum size=\myminsize] at (6,3) {\textcolor{white}{\citenum{paolillo2023dynamical}}};
    \node[circle, fill=dynamics, minimum size=\myminsize] at (7,3) {\textcolor{white}{\citenum{yang2023watch}}};
    \node[circle, fill=dynamics, minimum size=\myminsize] at (5,4) {\textcolor{white}{\citenum{yusuf2018playing}}};
    \node[circle, fill=dynamics, minimum size=\myminsize] at (6,4) {\textcolor{white}{\citenum{nair2017combining}}};
    \node[circle, fill=dynamics, minimum size=\myminsize] at (7,4) {\textcolor{white}{\citenum{pathak2018zeroshot}}};
    \node[circle, fill=dynamics, minimum size=\myminsize] at (5,5) {\textcolor{white}{\citenum{pavse2020ridm}}};
    \node[circle, fill=dynamics, minimum size=\myminsize] at (6,5) {\textcolor{white}{\citenum{guo2019hybrid}}};

    \node[circle, fill=hybrid, minimum size=\myminsize] at (7,5) {\textcolor{white}{\citenum{kidambi2021mobile}}};
    \node[circle, fill=hybrid, minimum size=\myminsize] at (5,6) {\textcolor{white}{\citenum{zhu2020opolo}}};
    \node[circle, fill=hybrid, minimum size=\myminsize] at (6,6) {\textcolor{white}{\citenum{yu2020giril}}};
    \node[circle, fill=hybrid, minimum size=\myminsize] at (7,6) {\textcolor{white}{\citenum{torabi2021dealio}}};
    \node[circle, fill=hybrid, minimum size=\myminsize] at (5,7) {\textcolor{white}{\citenum{gangwani2021imitation}}};
    \node[circle, fill=hybrid, minimum size=\myminsize] at (6,7) {\textcolor{white}{\citenum{wang2017robust}}};
    \node[circle, fill=hybrid, minimum size=\myminsize] at (7,7) {\textcolor{white}{\citenum{liu2018raw}}};
    \node[circle, fill=hybrid, minimum size=\myminsize] at (5,8) {\textcolor{white}{\citenum{baram2017endtoend}}};
    
    \node[rectangle, fill=white, align=left, text width=3cm, text height=4.5cm] at (9.5, 6.5) {%
        \begin{tikzpicture}
            \node[font=\bfseries] at (0, 3.6) {Our Taxonomy};
            \node[circle, fill=adversarial, minimum size=\myminsize] at (-1, 2.7) {};
            \node[right=0.2cm] at (-0.75, 2.7) {Adversarial};

            \node[circle, fill=bc, minimum size=\myminsize] at (-1, 1.8) {};
            \node[right=0.2cm] at (-0.75, 1.8) {BC};

            \node[circle, fill=dynamics, minimum size=\myminsize] at (-1, 0.9) {};
            \node[right=0.2cm] at (-0.75, 0.9) {Dynamics};

            \node[circle, fill=hybrid, minimum size=\myminsize] at (-1, 0) {};
            \node[right=0.2cm] at (-0.75, 0) {Hybrid};
        \end{tikzpicture}
    };
    
\end{tikzpicture}
    \caption{Comparison between taxonomies.  We remove online methods since both taxonomies are equal in that matter. }
    \label{fig:papers}
    \Description{Comparison between past taxonomies and current.}
\end{wrapfigure}
 their actions and, afterwards, using this model in the learning process of the policy.
Conversely, model-free methods do not use a model of the environment.
Instead, they learn by trial and error, similar to reinforcement learning approaches.
Model-based methods offer a trade-off in time and sample efficiency by leveraging the number of samples a teacher provides, the less time it takes to train a policy.
However, model-free methods are usually more robust to unseen states since they learn by trial and error, which gives them more access to diverse states (outside the teacher's dataset), resulting in more generalisation from their policies.
Lastly, imitation learning methods can also be classified as online.
Online imitation learning methods assume that we have access to the teacher's policy.
Since teacher policies are seldom available, these learning methods are less common.

Nevertheless, this classification can be quite coarse and does not truly reflect the nature of some methods.
Instead, we propose the taxonomy with the main five categories given in Fig.~\ref{fig:taxonomy}, where the numbers in parentheses indicate the total of papers surveyed in each category.
The key difference is that our classification distinguishes model-free methods that are based on behavioural cloning (Section~\ref{sec:sub:bc}) --- the simplest form of imitation learning --- from adversarial methods (Section~\ref{sec:sub:adv}); and partitions model-based methods into dynamics methods (Section~\ref{sec:sub:dmm}) and hybrid methods (Section~\ref{sec:sub:hybrid}).
For an illustration of how the IL papers discussed fit within our proposed taxonomy and within the traditional classification see Fig.~\ref{fig:papers}.

In our review, we found $50$ publications proposing new methods, which accounts for a big part of the reviewed work.
We expected a high percentage of work to be methodological papers since imitation learning borrows environments and metrics from the reinforcement learning literature.
On the one hand, this trend is sensible since there were fewer imitation learning methods in the past.
On the other hand, this focus on imitation learning methodology shows a lack of research on other subjects pertinent to imitation learning, such as resiliency~\cite{gavenski2022how} and evaluation processes~\cite{memmesheimer2023recognition,verma2023data}.
The most common approach was adversarial learning (following the taxonomy illustrated in Fig.~\ref{fig:taxonomy}) with $15$ published work, followed by $14$ dynamic, $11$ behavioural cloning, $8$ hybrid, and $2$ online methods.
Although we expected to find many methods that use dynamic models, an interesting trend is the use of hybrid approaches to create imitation learning agents.
We hypothesise that hybrid approaches and dynamic models are also becoming more popular with more researchers from backgrounds outside the reinforcement learning field.
Indeed, hybrid methods offer a better trade-off between efficiency and effectiveness, which is a reason for their increase in popularity in recent years.
Fig.~\ref{fig:papers} shows the difference between our taxonomy  (colours) and other surveys taxonomies (location) for each of the papers present in this work.

\subsection{Behavioural Cloning} \label{sec:sub:bc}

The earliest approach to imitation learning was Behavioural Cloning~\cite{pomerleau1988alvinn,michael1999bc}, which reduces the problem of learning to imitate a teacher into a supervised problem.
In it, the agent learns to predict the most likely action given a state $\argmax P(a \mid s_t)$ based on a teacher's dataset.
To be precise, the agent uses a state-action pair of demonstrations $(s_t, a_t)$ at time $t$ from a proficient source to learn to act as the source based on previously seen data.
Behavioural cloning tries to approximate the agent's \textit{trajectory} to that of its teacher.
A trajectory is a coherent sequence of demonstrations or experiences from one cycle of the teacher or agent's interaction with the environment.
Such an approach becomes costly for more complex scenarios, requiring a large number of samples and information about the action's effects on the environment. 
For example, the number of samples used to solve tasks involving a higher number of possible actions (e.g., continuous actions) or intricate dynamics is approximately a hundred times higher than for classic control tasks~\cite{torabi2018bco}.
This problem occurs for behavioural cloning policies since they fall short when approximating unseen states to known trajectories (generalisation), a common problem from supervised learning approaches.
Additionally, requiring demonstrations becomes costly due to the need for labelled pairs.
Usually, recording these trajectories involves training an agent via reinforcement learning, but some researchers record themselves playing, and as a result, they may not be demonstrating an optimal solution to the problem being solved. 
This scenario is not problematic as long as the measure of success depends on how good the agent is at imitating the teacher, rather than how good it is at actually solving the problem. 
Section~\ref{sec:metrics} details this point further.

Newer approaches usually employ behavioural cloning as a \textit{bootstrapping} mechanism.
Bootstrapping a policy means using a machine learning approach, such as supervised learning, to acquire knowledge from the environment or desired behaviour before applying another learning technique to fine-tune the agent's performance.
An example is the work of \citeauthor{lynch2020learning}~\cite{lynch2020learning}, where the authors use behavioural cloning to condition a policy to a set of goals and, afterwards, use planning to create an optimal trajectory.
A second possible bootstrapping application is the work of \citeauthor{daftry2017mav}~\cite{daftry2017mav}, in which the researchers record themselves flying a drone, apply behavioural cloning to learn their flying behaviour, and, as the final step, use reinforcement learning so the agent learns to adapt to different seasons (when the images look different).
These approaches display some of the benefits of using behavioural cloning.
The agent learns more efficiently \textit{offline} (without requiring direct environment access) by applying this technique and, afterwards, using the acquired knowledge to reduce the number of steps required for less efficient learning approaches, such as reinforcement learning.
Additionally, behavioural cloning coupled with bootstrapping can condition the policy to a more human-like behaviour (when humans provide the data).

Conversely, behaviour cloning remains the sole learning approach when the work mainly focuses on aspects outside the learning process.
One common approach among work such as~\cite{yan2017oneshot,chen2020cheating,zhang2018deep,carroll2019overcooked} is coupling behavioural cloning with other learning mechanisms, such as attention~\cite{yan2017oneshot} or domain-shift~\cite{chen2020cheating,zhang2018deep}.
We use these as mere examples of mechanisms applied to imitation learning.
Moreover, in these cases, the behavioural cloning approach could be swapped for another imitation learning paradigm with correct adaptations since the cost of acquiring labelled demonstrations remains significant.

\subsection{Dynamics Model methods} \label{sec:sub:dmm}
A possible approach for solving the task of mimicking a teacher without any direct action information is by employing the use of dynamics models.
These approaches are classified as model-based imitation learning methods since they learn a model of the environmental dynamics.
Generating dynamics models involves some form of \textit{online play}, which is when the agent interacts with the environment and focuses on a specific task.
Dynamic models can appear in two forms: \textit{inverse} and \textit{forward}.

\subsubsection{Inverse Dynamics Models} \label{sec:sub:sub:idm}

Inverse Dynamics Models avoid the need for labelled pairs in behavioural cloning by encoding environmental physics and retrieving the likelihood of each action given a state transition $P(a \mid s_t, s_{t+1})$. 
\citeauthor{nair2017combining}~\cite{nair2017combining} use this form of learning to teach a robot to manipulate a rope in different ways.
In their work, the dynamics model receives the current state of the environment and a state from a sequence of human demonstrators as the goal state from that sequence.
With both states, the model predicts the action responsible for the desired transition conditioned to a desired goal.
This process is applied sequentially, and the authors perform all experiments using only matrices (image) states.

\citeauthor{torabi2018bco}~\cite{torabi2018bco} later implement an inverse dynamics model, which they use in vector states for control and robotic tasks and coined as \textit{Behavioural Cloning from Observation}.
It uses its randomly initialised policy to learn a mapping function without access to the teacher's action by creating a dataset containing labelled actions.
Hence, \citeauthor{torabi2018bco}'s method does not rely on a desired state (or goal) from its teacher.
Their method uses demonstrations $(s_t, a_t, s_{t+1})$ from $\pi_\theta$ to learn the dynamics model.
This dynamics model learns a uniform transition function (given the uniformly distributed dataset due to the random initialisation from policy $\pi$ with weights $\theta$) and creates self-supervised labels for the teacher's observations.
Provided with these labelled samples, \citeauthor{torabi2018bco}'s work trains its policy in a supervised manner using behavioural cloning.
Following \citeauthor{torabi2018bco}'s work, other methods augmented their approach to improve its performance~\cite{monteiro2023self}, stability~\cite{gavenski2020iupe}, sample efficiency~\cite{pavse2020ridm,gavenski2021self}, and to other domains~\cite{yang2023watch,raychaudhuri2021cross}.

Most notably, our work~\cite{gavenski2020iupe} improves \citeauthor{torabi2018bco}'s work by applying an exploration mechanism and fine-tuning its sampling mechanism, the first imitation learning method to use an exploration mechanism.
The method assumes that the dynamics and policy models are not always sure about the correct action and samples from each model's output using a softmax distribution.
Therefore, if a model has an equal distribution between two actions, it will select each action $50\%$ of the time.
Conversely, if a model's output mainly favours an action, such as $[2.5, 0.1]$, the model will choose the first action approximately $90\%$ of the time.

As behavioural cloning methods, inverse dynamics methods also combine their approach with reinforced learning methods~\cite{pavse2020ridm,gavenski2021self}.
\citeauthor{pavse2020ridm}'s work involves two phases.
In the first phase, an inverse dynamics model is randomly initialised and learns state transitions from a random policy, just as \citeauthor{torabi2018bco}'s work would.
In its second phase, the method alternates between generating agent experiences (with environment rewards) and optimising the policy with teacher demonstrations.
This two-phase procedure aims to find the optimal policy in terms of total task reward (which may outperform the teacher) by using the expert demonstration as a guide.
Our work~\cite{gavenski2021self} similarly alternates reinforcement learning with imitation learning.
The difference between these approaches is that instead of using all experiences, the second applies the goal-aware sampling mechanism from \citeauthor{gavenski2020iupe}'s work and limits the size of the replay buffer from its reinforced learning counterpart to have less drastic updates when learning with experiences.

\subsubsection{Forward Dynamics Models} \label{sec:sub:sub:fdm}

Like inverse dynamics models, forward dynamics models model the environment's dynamics using state transitions.
These models predict the next state, given some conditioning.
Most researchers condition these models' predictions with the environment's actions to generate the next state~\cite{pathak2018zeroshot,edwards2019ilpo}.
However, others rely on temporal information, such as using all states until the current moment~\cite{paolillo2023dynamical,rodrigo2023stable}.

Most notably, \citeauthor{pathak2018zeroshot} use the same idea from \citeauthor{nair2017combining}, where it conditions its policy with a goal.
However, instead of using current and goal states to predict the actions, the authors use current and goal state features coupled with the last action to predict the most likely next action and use the current and the predicted action to generate the next state.
\citeauthor{pathak2018zeroshot}'s policy uses a common technique to predict its information: using all previous states until timestep $t$, which helps to create a consistent action considering all previous states.

Nevertheless, the action information might be unavailable or can be insufficient.
For example, suppose an agent walks towards an impassable wall and hits it.
Even though the agents performed an action, such as `move forward', the final transitions would show no movement.
Thus, the agent should consider that the action responsible for such a transition is `no action', which is sometimes absent in the environment dynamics.
\citeauthor{edwards2019ilpo}'s work applies this premise to forward dynamics models by using these \textit{latent actions} to condition its next state.
By doing so, \citeauthor{edwards2019ilpo}'s method creates more faithful transitions according to the environment dynamics.
As a final step, the method requires remapping the learned actions into environmental actions.

Forward dynamics models are less popular than inverse dynamics models, given their nature of predicting the next state from the Markov Decision Process, which is more challenging than predicting the action responsible for a state transition~\cite{ferrolho2021inverse}.
The agent can lower its generalisation capabilities by adding this sequential information into a behavioural cloning technique when the trajectory deviates from its teacher's dataset.

\subsection{Adversarial learning methods} \label{sec:sub:adv}

Adversarial learning and inverse reinforcement learning methods share the same task in imitation learning. 
Instead of trying to reproduce the teacher's behaviour by applying some form of supervised learning, these methods create an artificial reward function, which conditions the agent by rewarding similar behaviours.
By creating this artificial reward function, imitation learning approaches can use reinforcement learning optimisation techniques to learn policies with similar behaviours from demonstrations.
Most work use adversarial learning methods to create these reward functions~\cite{stadie2017third,torabi2019video,jiaming2018multiagent,yunzhu2017infogail,hu2023safe,peng2018variational,wen2019efficient,yin2022planning,torabi2018gaifo,chang2022ilflow}; however, some earlier work rely on other inverse reinforcement learning techniques~\cite{finn2016guided,lee2014mario,shih2022conditional}.
This method is model-free since the policy freely acts in the environment and learns by trial and error without modelling the environment dynamics.
Nevertheless, not all inverse reinforcement learning methods use online play to learn its policy.

Before applying generative models, inverse reinforcement learning in an imitation learning setting used demonstrations to guide the agent's learning by applying a distance-based optimisation function to the teacher's and agent's trajectories.
\citeauthor{finn2016guided}~\cite{finn2016guided}'s work uses inverse entropy to retrieve an artificial reward function based on a random controller and the teacher's demonstrations.
With the artificial reward, their method optimises a policy function, creating new demonstrations and allowing for further refinement of the retrieved reward~function.

\citeauthor{ho2016generative}~uses \textit{adversarial learning}~\cite{goodfellow2014gan} and maximum entropy inverse reinforcement learning~\cite{ziebart2008maximum}.
Adversarial learning is a type of training where a model trains to generate data according to a dataset, and a discriminator model has to learn how to discriminate samples from the dataset and those generated by the first model.
The method uses a generative model, the same one that forward dynamics models use, to generate trajectories and a discriminator model to discriminate between teachers' and students' trajectories.
Therefore, the policy role in this setting is to `fool' the discriminator model by acting accordingly to the teacher demonstrations.
However, the method presents two problems:
\begin{enumerate*}[label=(\roman*)]
    \item it assumes that the teacher's actions are available during training, which inherits the cost of behavioural cloning approaches; and
    \item it is susceptible to local minima during its optimisation process, which requires prolonged environmental interactions.
\end{enumerate*}
Nevertheless, \citeauthor{ho2016generative} is more sample-efficient than behavioural cloning, which decreases the severity of the first point.
\citeauthor{ho2016generative}'s work~\cite{ho2016generative} is the most known work and perhaps one of the most influential, with several methods following the same setting~\cite{yunzhu2017infogail,torabi2018gaifo,peng2018variational,monteiro2023self}.
Most notably, 
\citeauthor{yunzhu2017infogail}'s work~\cite{yunzhu2017infogail} assumes that a dataset consists of various behaviours from one or more teachers.
So, instead of creating a policy that tries to predict the most likely action solely on the state alone, it also conditions the agent with a teacher's parameter.
Additionally, \citeauthor{peng2018variational}'s work changes how the discriminator discriminates between teacher and student.
In it, the authors constrain the amount of information given to the discriminator by applying the output of the generative model into an encoder.
Therefore, the discriminator uses an encoded value instead of the state to discriminate.
\citeauthor{peng2018variational} show that applying this bottleneck helps the policy to generalise better and achieve better results than other adversarial methods that rely on state information.

\citeauthor{torabi2018gaifo}'s work~\cite{torabi2018gaifo} applies a similar strategy to \citeauthor{ho2016generative}'s work to apply imitation learning from observation.
It assumes that adversarial methods follow a convex conjugate optimisation problem, where the generative model and policy function learn to approximate state transitions alone.
To approximate teacher and student, a discriminator has to discriminate over the agent and teacher trajectories observations.
By learning to differentiate between origins, the discriminator retrieves an artificial reward function, which the method uses to optimise the policy. 
By applying these changes, \citeauthor{torabi2018gaifo}'s work improves adversarial methods performance in more dynamic environments and reduces the cost of learning new policies.
\citeauthor{torabi2018gaifo}'s work has also seen further improvement by \citeauthor{wen2019efficient}'s work~\cite{wen2019efficient}.
\citeauthor{wen2019efficient} use an adversarial learning approach to create a policy that minimises the policy with a no-regret function bounded to one.
Therefore, if the optimisation function is zero, the policy acts as the teacher; however, if the artificial reward function returns one, the policy is as far as possible from its teachers.
As with other work, acquiring new demonstrations to further policy improvement is done via online play, which becomes costly if many interactions are needed.
Additionally, \citeauthor{torabi2019video} further explore his methods capabilities by applying it to vision tasks~\cite{torabi2019video} by using convolutional neural networks as the discriminator model.
Since vectorised information is not typical for applications, optimising a policy using only visual information from a convolutional neural network while holding the vectorised representation as the input for its policy allows for more optimal solutions.

Although adversarial learning methods became the standard for model-free approaches, inverse reinforcement learning techniques have also demonstrated remarkable advancements recently.
\citeauthor{chang2022ilflow}'s work~\cite{chang2022ilflow} discards the usage of discriminator models to match the policy's trajectories to the teacher's using a distance metric.
However, using this approach, the method can enter a seeking behaviour state, producing low probabilities for observations from the teacher data.
Therefore, \citeauthor{chang2022ilflow} implement a noise-conditioned normalising flow, which adds noise to observations conditioned to the state during training and nothing during evaluations.
Doing so reduces the chance of low probabilities and biased predictions from the policy.
In multi-agent systems, inverse reinforcement learning methods have also achieved some success.
In \citeauthor{jiaming2018multiagent}'s work~\cite{jiaming2018multiagent}, the method is divided into centralised (or cooperative), decentralised, and zero-sum.
In centralised and decentralised approaches, the policy operates with the environment to match the teacher reward by using adversarial techniques. 
On the other hand, in the last approach, the policy interacts directly with the discriminator to match the reward from the teacher.
Conditioning also appears in multi-agent settings.
For example, \citeauthor{shih2022conditional} proposes to create a policy that adapts its strategy depending on other agents~\cite{shih2022conditional}.
Therefore, the method has to condition itself by classifying which strategy from the teachers' dataset is being used.

Finally, these approaches also appear in different settings and tasks.
For example, \citeauthor{shang2021self} assumes that the agent only has access to third-person views from the teacher (a more human-like experience of observing others)~\cite{shang2021self}.
Thus, outside of learning the teacher's behaviour, the agent must also learn to adapt the data from its teacher to suit its first-person view when enacting the same task.
In \citeauthor{yin2022planning}'s work~\cite{yin2022planning}, the method has to predict a sequence of actions from a behavioural cloning and adversarial agent. 
After both policies predict all actions, a planning model forms a plan of which actions should be used throughout the trajectory.
Finally, in the domain of self-driving vehicles, \citeauthor{hu2023safe}'s work~\cite{hu2023safe} experiments with sharing weights with the policy and generative model, which the method uses to optimise the policy using a reinforcement learning technique.

\subsection{Hybrid approaches} \label{sec:sub:hybrid}

Hybrid methods leverage dynamics models with adversarial techniques to learn their policies.
Recent work in imitation learning has focused on taking inspiration from adversarial and dynamics methods~\cite{zhu2020opolo,kidambi2021mobile,yu2020giril,torabi2021dealio,monteiro2023self,liu2018raw,baram2017endtoend,wang2017robust,gangwani2021imitation}.
By combining dynamics models and adversarial learning, these methods create more robust policies since they have a temporal signal that helps minimise teacher and student differences and some intrinsic knowledge about the environment dynamics.
Although they use discriminators to achieve their task, these methods have significantly improved imitation learning efficiency, with some requiring a single trajectory~\cite{zhu2020opolo}.

Towards sample efficiency and the issue of saddle points, \citeauthor{wang2017robust}~\cite{wang2017robust} use a variational autoencoder to increase diversity with relatively fewer demonstrations and achieve one-shot learning to new trajectories. 
They apply a temporal neural network to create a non-stationary policy.
It is worth noting that even though the generative model relies on a forward dynamics model, the agent's optimisation does not rely on it, increasing efficiency during optimisation.

In learning from demonstration hybrid methods, \citeauthor{baram2017endtoend} develop a method that optimises the forward dynamics model~\cite{baram2017endtoend}.
Instead of concatenating state and action into a neural network to predict the next state, the method uses a recurrent neural network (temporal) to encode the state and another non-temporal encoder and perform an element-wise multiplication to feed as the input of a third neural network.
By doing so, the authors achieve teacher behaviours earlier during training and reduce the number of teacher samples.
In the same context, hybrid methods have also seen use in partial demonstrations from teachers~\cite{yu2020giril}.
In their work, the demonstrator is a single teacher's life in Atari environments (less than a whole episode -- three lives), and \citeauthor{yu2020giril} use a variational autoencoder to encode a forward dynamics model and a discriminator to learn an artificial reward function.
Although learning from demonstration work is not as common as its observation counterpart, these methods tackle problems significant for imitation learning agents.
Removing the complexity of learning policies without action information allows researchers to tackle other, more complex problems.

Hybrid methods for learning from observation are more prevalent in recent research.
These methods focus on efficiency~\cite{zhu2020opolo,kidambi2021mobile,torabi2021dealio,monteiro2023self} and eliminating the seeking behaviour~\cite{zhu2020opolo,liu2018raw}.
\citeauthor{zhu2020opolo}'s work~\cite{zhu2020opolo} solves the issue of imitation learning agents adopting a seeking mode by using a discriminator to swap between a covering and seeking mode when the agent heavily diverges from the teacher.
It does so by using ten teacher's trajectories for each environment, but \citeauthor{zhu2020opolo} assume that no more than a single state-action pair can result in the same next state, which is not valid in most environments where there is momentum for example. 
Most notably, \citeauthor{kidambi2021mobile}'s work uses a forward dynamics model with adversarial learning and couples it with an exploration mechanism~\cite{kidambi2021mobile} with ten teacher trajectories.
However, their method assumes that all episodes initiate in the same state, which causes it to fail to generalise as state sizes grow.
Moreover, \citeauthor{kidambi2021mobile}'s method exploration mechanism comes with the cost of using an ensemble method (more than one model during prediction), which later work by \citeauthor{monteiro2023self}'s solves.
In addition to solving efficiency problems, \citeauthor{liu2018raw}'s work~\cite{liu2018raw}, the authors experiment with encoders across different contexts, creating dynamics models to translate between teacher and student states and apply a discriminator model to approximate both behaviours.
By doing so, their work eliminates the need to fine-tune a model every time an agent's context changes and only requires training a new encoder, allowing for translating teacher behaviours into new contexts.

More recently, \citeauthor{gangwani2021imitation}'s work assumes that a policy has to learn from a teacher's trajectory that diverges from the agent.
Different from \citeauthor{liu2018raw}'s work, teachers and agents are in the same context, but the transition function for both is different.
In other words, the teacher's observations should guide the policy towards a goal, similar to other methods that condition their policies as a goal state.
In their work, the policy learns a forward dynamics model to output the next state from the agent's transition function perspective that aligns with the teacher's trajectories.
\citeauthor{gangwani2021imitation}'s work~\cite{gangwani2021imitation} shows that it performs better in these settings than other methods. 
However, the results are far from the teacher, which shows that such a setting needs additional~research.

We note that hybrid methods have become more popular.
Although adversarial learning methods are less sample efficient, they help model-based methods generalise and learn more human-like behaviour.
Methods such as \citeauthor{kidambi2021mobile}'s and \citeauthor{zhu2020opolo}'s work achieve teacher results with ten or fewer trajectories, but some recent studies show that there are still other mechanisms to research.
For example, exploration mechanisms and cross-domain contexts have been uncommon in imitation learning approaches.
By exploring, agents can acquire knowledge unattainable before due to the need for more variety in teacher samples.
When applying cross-domain contexts, imitation learning agents can learn from a third-person view or different settings for the agent, such as fewer joints in a robot.

\subsection{Online Imitation Learning} \label{sec:sub:online}
A different approach to learning from a teacher is online imitation learning.
In this approach, the agent must also learn the teacher's behaviours. 
However, instead of using demonstrations from a pre-recorded dataset, online imitation learning involves access to the teacher's information in real-time.
Online learning can be detrimental when designing a reward function that is challenging to define or a policy function that is difficult to model.
Given the requirement of direct access to a teacher, such as another agent, this approach is less commonly researched than the other areas.
Nevertheless, online imitation learning is more suitable when real-time interaction with a teacher is possible or when the agent needs to adapt to changing conditions.

\citeauthor{ross2011reduction} implement an iterative online imitation learning algorithm that uses demonstrations from online play and teachers to learn a behaviour~\cite{ross2011reduction}.
It uses a \textit{Follow-The-Leader} technique to optimise the policy behaviour based on the demonstrations it retrieves during the first iterations and from the teacher.
Follow-The-Leader tries to minimise a cumulative loss function or maximise the agent's reward sequentially.
Since reward values are not available to the agent in imitation learning approaches, Follow-The-Leader tries to minimise the difference between teacher and student behaviour.
\citeauthor{lavington2022online} further improved the Follow-The-Leader strategy by applying a regularisation term, which they coined as Follow-The-Regularised-Leader~\cite{lavington2022online}.
In it, \citeauthor{lavington2022online} consider that the teacher's demonstrations might have noise, and the Follow-The-Leader approach can lead to oscillations resulting in non-optimal behaviour.
The researchers show that a policy achieves better results after applying regularisation to the optimisation function.

\subsection{Learning Methods Discussion}\label{sec:learning_conclusion}
Upon analysing the literature from this section, we notice a change in the development of new learning methods, which use hybrid mechanisms.
Instead of relying solely on dynamics models or adversarial approaches to learn the teacher's behaviour, these methods mix different aspects from both sides to achieve better results.
Furthermore, recent approaches focus more on imitation learning from observation and efficiency problems.
In Section~\ref{sec:reflections}, we further expand on challenges researchers may focus on rather than efficiency and also reflect on different aspects learning methods may not account for.

This literature review shows a lack of imitation learning methods applied to online learning and multi-agent systems.
We believe this dearth of work for online learning comes from the premise of having access to the teacher, which limits the number of applications and the complexity of learning from a policy in an environment where more than one agent impacts the transition function.
Like online learning, multi-agent systems use highly dynamic environments to learn their policies.
In these environments, the overall combination of all possible states can be virtually infinite, which leads to imitation learning agents requiring a lot of data or drastic assumptions on how to model the environment (that might not hold true in diverse cases).
Therefore, we argue that imitation learning methods tailored specifically to the challenges posed by these applications need to be researched and~developed.
\section{Imitation Learning Environments} \label{sec:environments}

Environments are a vital part of the imitation learning process.
They simulate various tasks and domains and allow the agent to interact with the simulation as they would in a real-world application.
Creating an environment most often involves creating the environment's rules, such as physics for each object, and implementing rules to guide the agent, for example, how agents should interact.
Nevertheless, they allow different work to evaluate their results using the same dynamics, creating a more fair comparison since all methods receive the same demonstrations and solve the same problem.
In the literature review, we notice a lack of consensus on which environments imitation learning literature should use.

In the literature on agents, there are different descriptions of what environments consist of.
We follow \citeauthor{russell2020artificial}'s definition, which describes environments in four different aspects~\cite[Chapter. 2]{russell2020artificial}.
The first aspect entails the \textit{accessibility} of the states from the environment.
Most commonly, environments use fully observable states, meaning the agent can obtain complete, accurate, up-to-date information. However, most real-world environments are inaccessible in this sense.
The second aspect is certainty.
\textit{Deterministic} environments are those in which any action has a single guaranteed effect.
In these environments, there is no uncertainty about the state resulting from an action.
We call environments that are not deterministic \textit{non-deterministic}.
The third aspect describes the environment dynamics.
\textit{Static} environments are those that are safe to assume will remain unchanged except by the performance of an action by an agent.
In contrast, a \textit{dynamic} environment has other processes operating, changing in ways beyond the agent's control, such as the physical world, which is a highly dynamic environment.
Lastly, the fourth aspect entails how things are represented.
\textit{Discrete} environments are those with fixed, natural numbers and a finite set of actions, while \textit{continuous} are those represented by rational numbers, hence an infinite set of actions with upper and lower limits.

\begin{figure*}[b]
    \centering
    \begin{minipage}{0.5\linewidth}
        \centering
        \begin{tikzpicture}[
    node distance=3cm,
    every node/.style = {
        rectangle, 
        rounded corners,
        minimum height=1cm,
        font=\Large \bfseries,
        text centered
    },
    node/.style = {
        minimum width = 5cm
    },
    subnode/.style = {
        minimum width = 4cm
    },
    arrow/.style = {
        ultra thick,
        ->,
        >=latex
    }
]

    \node (root) [%
        node, fill=white!10, draw=black, minimum width={width("Imitation Learning Environments (66)")+100pt}
    ] {Imitation Learning Environments (66)};

    \node (precision) [node, below of=root, fill=cb_blue!25, draw=cb_blue!75] {Precision (10)};
    \node (prec_domain) [%
        subnode, 
        below of=precision, 
        fill=cb_blue!25, 
        draw=cb_blue!90, 
        node distance=1.5cm, 
        xshift=1cm,
        align=center
    ] {Domain (7)};
    \node (prec_task) [%
        subnode, 
        below=of prec_domain.west, 
        fill=cb_blue!25, 
        draw=cb_blue!75, 
        yshift=1.5cm,
        anchor=west,
        align=center
    ] {Task (3)};

    \node (validation) [node, left of=precision, xshift=-3cm, fill=cb_orange!10, draw=cb_orange!50] {Validation (44)};
    \node (val_domain) [%
        subnode, 
        below of=validation, 
        fill=cb_orange!10, 
        draw=cb_orange!50, 
        node distance=1.5cm, 
        xshift=1cm,
        align=center
    ] {Domain (2)};
    \node (val_task) [%
        subnode, 
        below=of val_domain.west,
        fill=cb_orange!10,
        draw=cb_orange!50,
        yshift=1.5cm,
        anchor=west,
        align=center
    ] {Task (42)};

    \node (sequential) [node, right of=precision, xshift=3cm, fill=cb_purple!10, draw=cb_purple!50] {Sequential (12)};
    \node (seq_domain) [%
        subnode, 
        below of=sequential, 
        fill=cb_purple!10, 
        draw=cb_purple!50, 
        node distance=1.5cm, 
        xshift=1cm,
        align=center
    ] {Domain (5)};
    \node (seq_task) [%
        subnode, 
        below=of seq_domain.west,
        fill=cb_purple!10,
        draw=cb_purple!50,
        yshift=1.5cm,
        anchor=west,
        align=center
    ] {Task (7)};

    \draw [arrow]   (root) -- (precision);
    \draw [arrow]   ++(0, -1.25cm) -|
                    (validation.north);
    \draw [arrow]   ++(0, -1.25cm) -|
                    (sequential.north);

    \draw [arrow]   ([xshift=-1.5cm]precision.south) |- (prec_domain.west);
    \draw [arrow]   ([xshift=-1.5cm]precision.south) |- (prec_task.west);

    \draw [arrow]   ([xshift=-1.5cm]validation.south) |- (val_domain.west);
    \draw [arrow]   ([xshift=-1.5cm]validation.south) |- (val_task.west);

    \draw [arrow]   ([xshift=-1.5cm]sequential.south) |- (seq_domain.west);
    \draw [arrow]   ([xshift=-1.5cm]sequential.south) |- (seq_task.west);

\end{tikzpicture}
        \caption{Imitation learning environments taxonomy.}
        \label{fig:environments:taxonomy}
    \end{minipage}%
    \hfill
    \begin{minipage}{0.49\linewidth}
    \nextfloat
    \begin{subfigure}[b]{0.32\linewidth}
        \centering
        \includegraphics[width=\textwidth]{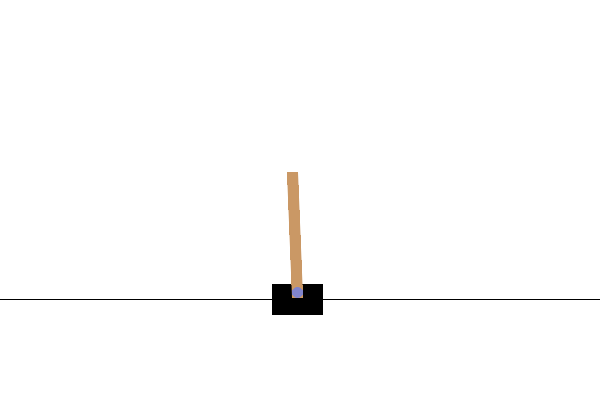}
        \caption{CartPole~\cite{barto1983neuronlike}}
        \label{fig:sub:cartpole}
    \end{subfigure}%
    \hfill
    \begin{subfigure}[b]{0.35\linewidth}
        \centering
        \includegraphics[height=2.2cm]{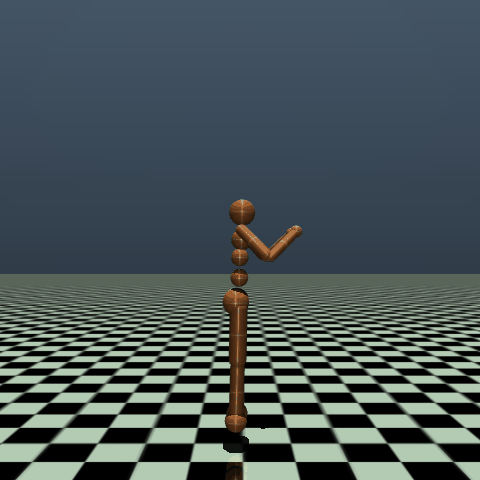}
        \caption{Humanoid~\cite{tassa2018deepmind}}
        \label{fig:sub:humanoid}
    \end{subfigure}%
    \hfill
    \begin{subfigure}[b]{0.3\linewidth}
        \centering
        \includegraphics[height=2.2cm]{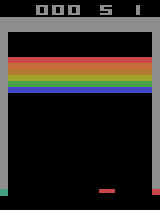}
        \caption{Breakout~\cite{bellemare2013atari}}
        \label{fig:sub:breakout}
    \end{subfigure}%
    \caption{Examples of validation environments.}
    \label{fig:validation}
    \Description{Examples of validation environments.}
    \end{minipage}%
\end{figure*}

Environments can also be domain-specific, such as autonomous driving, or more general, such as robotic simulations, which use 2D and 3D simulations to evaluate agents in various similar tasks.
Most work use the application to differentiate each environment, such as assistive and humanoid robotics~\cite{hussein2017imitation}.
However, such a classification is shallow since it only focuses on a specific set of problems per environment, not what the environment tests the agent on.
Therefore, we propose a threefold classification for these environments:
\begin{enumerate*}[label=(\roman*)]
    \item validation, which helps researchers to validate ideas without added complexity;
    \item precision, which helps to test agents that require less temporal modeling, but require precision; and
    \item sequential, which tests an agent's actions and their temporal consequences.
\end{enumerate*}
Fig.~\ref{fig:environments:taxonomy} presents the proposed taxonomy for the environments.


It is important to note that other forms of classifying environments exist, such as how the states are represented (images or vectors), actions (discrete or continuous), and whether a task is a maintenance task (a set of states are unreachable to the agent) or an achievement task (the agent should reach a set of states).
However, our classification focuses more on the agent's intent, such as `to be as precise as possible', than on the task or domain itself.
Additionally, this taxonomy also contains domain-specific and task-specific classifications after the broader term.
For each environment, we present whether the environment presents domain-specific or task-specific problems, which should help researchers decide if an environment is desirable for their experiments.

\subsection{Validation environments} \label{sub:validation}

Validation environments are more straightforward simulations in general.
These environments help to validate whether an agent can learn to perform a task from a teacher with no extra complexity involved.
Usually, the agent is only required to learn the task, and the environment does not diverge much when initialising the agent.
These environments are more commonly split into discrete and continuous action environments.

The most common discrete environment in the imitation learning literature is CartPole~\cite{barto1983neuronlike} (Fig.~\ref{fig:sub:cartpole}).
This environment is a perfect example of a validation environment.
It consists of a pole attached by an unactuated joint to a cart, which moves along a frictionless track.
The pole is placed upright on the cart, and the goal is to balance the pole by applying forces to the left and right of the cart.
Thus, CartPole is a maintenance task-specific environment.
Each episode starts with uniformly random values between $-0.05$ and $0.05$ for the cart position, cart velocity, pole angle and pole angular velocity. 
Therefore, each episode has a small variety, and since there is no differentiation between training and validation, such as different mechanics, the agent is not required to generalise.
Achieving an optimal result within the CartPole environment does not demonstrate that the agent can learn and perform complex tasks.
However, it is helpful as an initial validation.
Alongside the CartPole, the most common environments for discrete actions are the MountainCar~\cite{moore1990efficient} environment and Atari games~\cite{bellemare2013atari}.
In the MountainCar environment, a car is randomly placed at the bottom of a valley shaped like a sine wave. 
The only available actions are accelerating the car in either direction or not accelerating.
The MountainCar problem deals with momentum, making it harder for simple or random agents to solve since it carries some sequential impact.
If an agent builds momentum and selects to use force opposite to its direction, the agent will lower its speed and fail to reach the goal.
On the other hand, Atari games are environments that consist of visual state representations. 
They are commonly used in imitation learning tasks because they allow for visual qualitative assessment of the agent's behaviour.
Moreover, Atari games were a long benchmark for comparing how agents perform in different tasks and allow for a direct comparison with humans using the video game score~\cite{machado18arcade}.

The most common continuous environments are simulations from Multi-Joint dynamics with Contact (MuJoCo)~\cite{todorov2012mujoco} (Fig.~\ref{fig:sub:humanoid}).
In these environments, the agent is a robotic simulation of different joints and body parts configurations, such as a cheetah or a human.
The agent aims to move right by applying force to these joints and body parts.
Similar to the CartPole environment, these environments are simplistic in the sense of not posing new challenges to the agent.
The agent has no barrier to overcome or other dynamics it has to adapt to.
However, differently from the CartPole environment, these environments still pose a challenge in the sense that the agent has to learn to control its body joints together to achieve the goal.

Most work use a combination of validation and another environment type to experiment with the limitations of imitation learning agents.
However, some work~\cite{monteiro2020abco,torabi2018bco,torabi2018gaifo,torabi2021dealio,yin2022planning,yu2020giril,wen2019efficient} use only validation environments, which can be troublesome.
By only using these environments, the agent can learn to overimitate its teacher without comprehending the underlying task.
Additionally, considering the initialisation is not diverse, the agent can force a teacher's trajectory into the environment, causing the same problem as reward-based metrics.
In other words, during an agent's first interactions with an environment, the agent can predict actions that will take it to the same states present in the teacher's dataset, and afterwards, the agent will only have to perform state matching to achieve optimal results.
Other frameworks for simulation, such as Google's suite~\cite{tunyasuvunakool2020}, promise a diverse set of initialisations, but further testing shows that these simulations are vulnerable to the same seeking behaviour.
Table~\ref{tab:env_validation} shows all validation environments with \citeauthor{norvig2002modern}'s environment attributes information.

\subsection{Precision environments}

\begin{figure*}[b]
    \centering
    \begin{subfigure}[b]{0.24\textwidth}
        \centering
        \includegraphics[height=2.2cm, width=\textwidth]{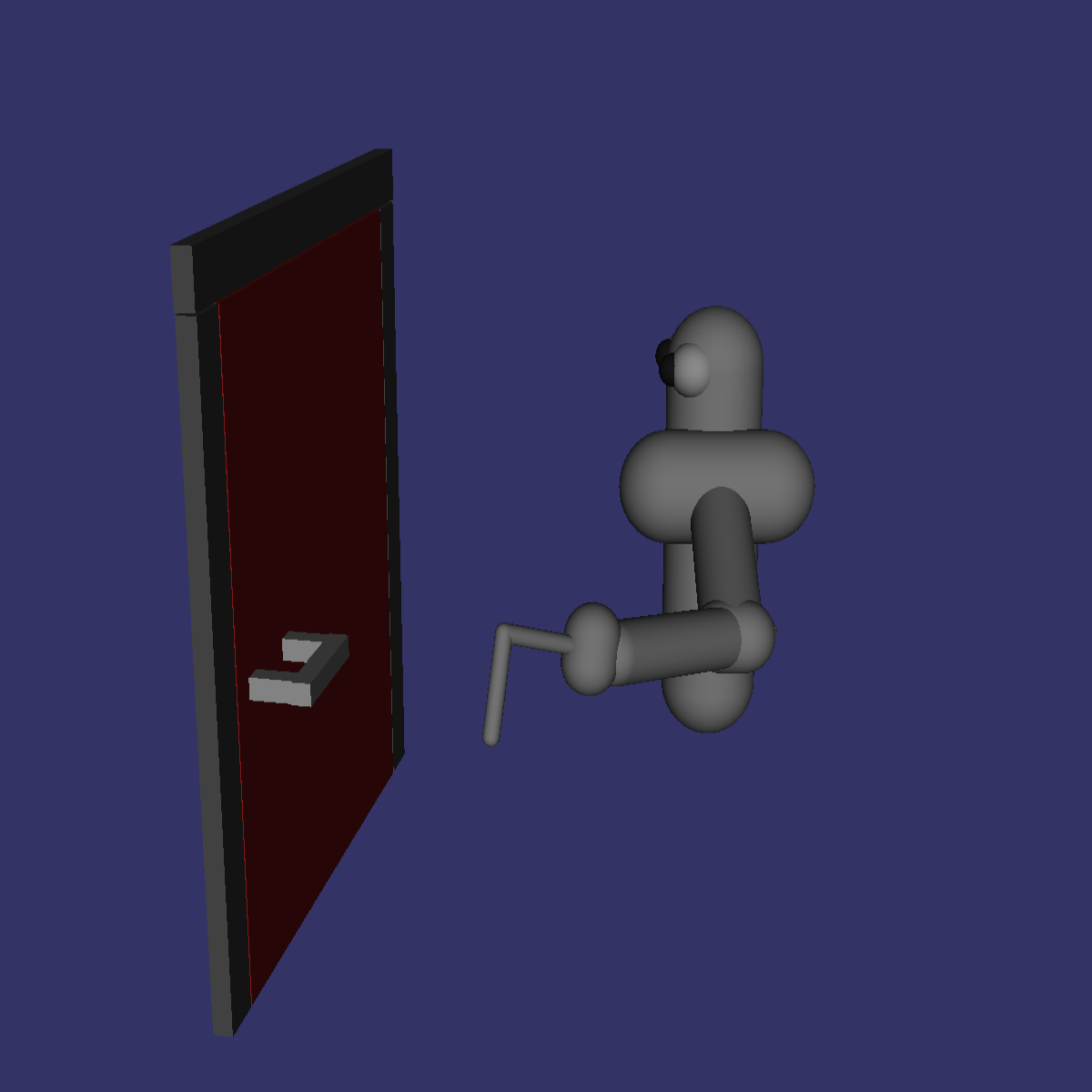}
        \caption{MuJoCo~\cite{todorov2012mujoco}}
        \label{fig:sub:mujoco}
    \end{subfigure}
    \hfill
    \begin{subfigure}[b]{0.24\textwidth}
        \centering
        \includegraphics[height=2.2cm, width=\textwidth]{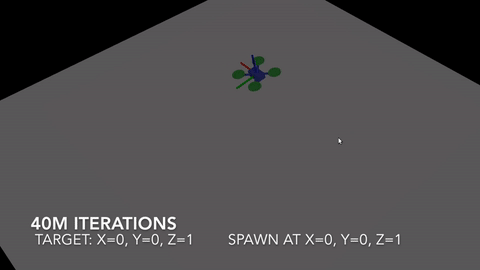}
        \caption{MAV~\cite{daftry2017mav}}
        \label{fig:sub:mav}
    \end{subfigure}
    \hfill
    \begin{subfigure}[b]{0.24\textwidth}
        \centering
        \includegraphics[height=2.2cm]{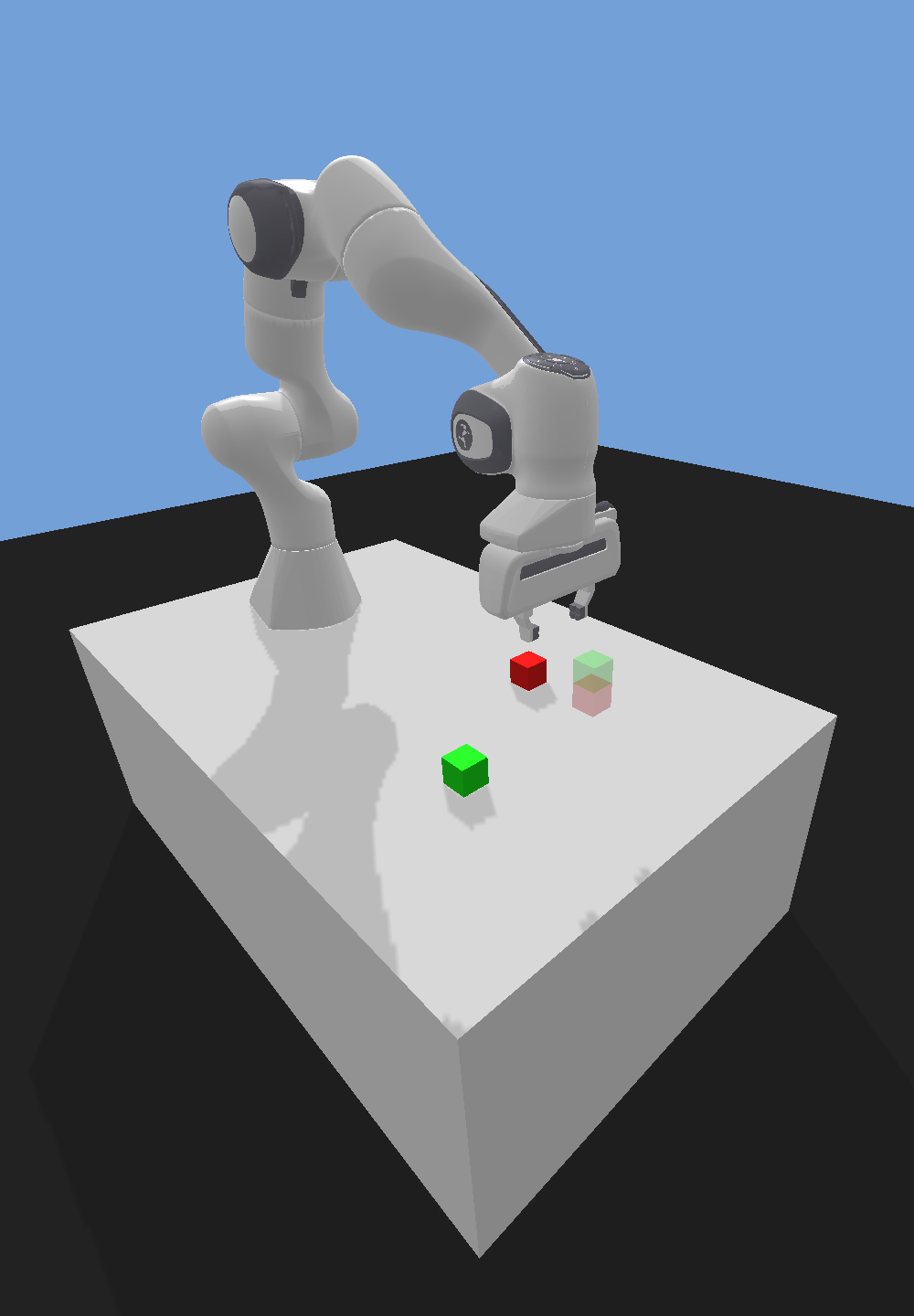}
        \caption{Panda-gym~\cite{gallouedec2021panda}}
        \label{fig:sub:pandas}
    \end{subfigure}
    \hfill
    \begin{subfigure}[b]{0.24\textwidth}
        \centering
        \includegraphics[height=2.2cm, width=\textwidth]{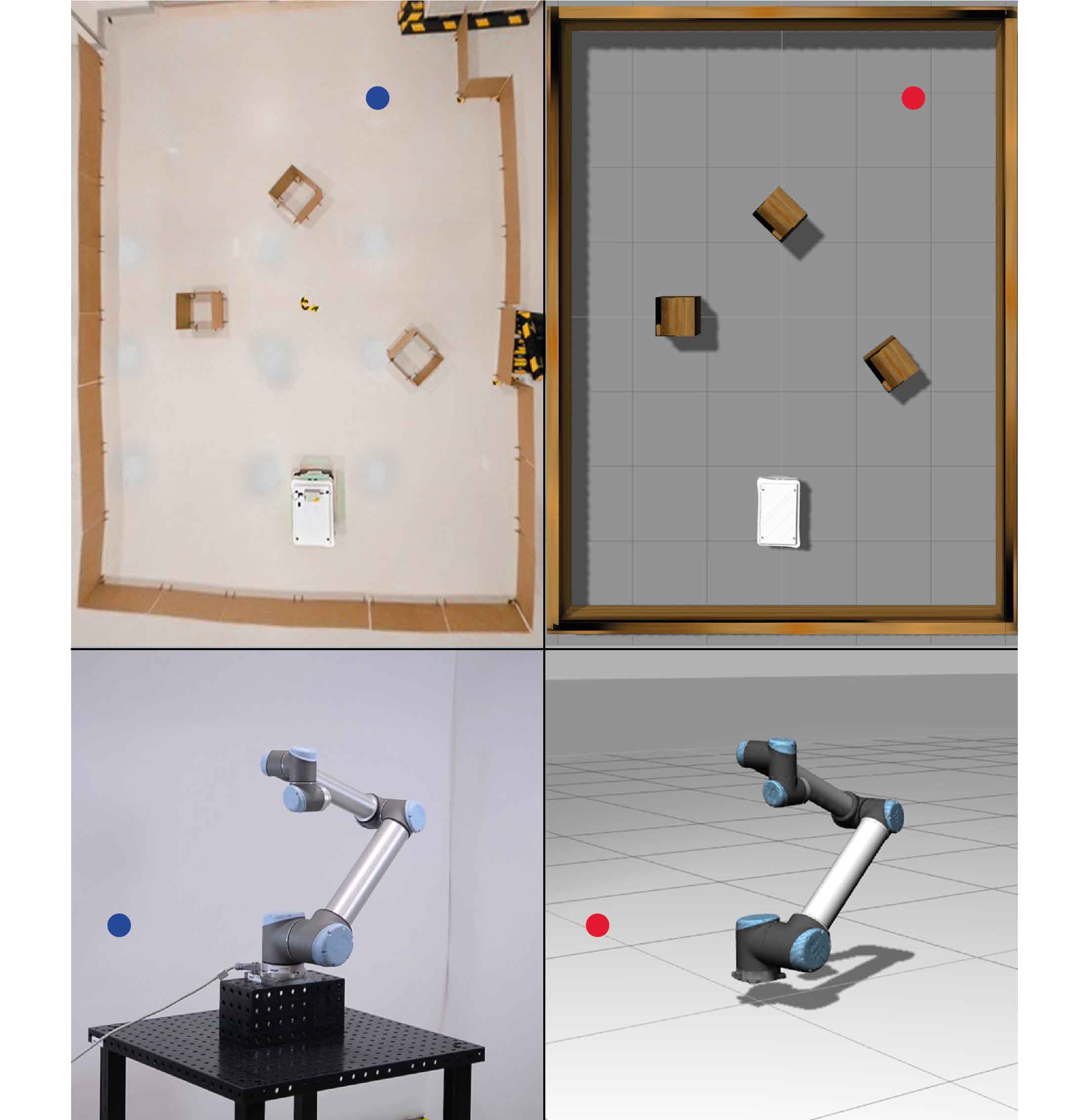}
        \caption{Robo-gym~\cite{lucchi2007robogym}}
        \label{fig:sub:robo}
    \end{subfigure}
    \caption{Examples of precision environments.}
    \Description{Examples of precision environments.}
    \label{fig:precision}
\end{figure*}

Precision environments are generally independent of long action consequences and require precision, as the name suggests, from the agent.
\citeauthor{russell2020artificial} classify this autonomy from time as episodic environments~\cite[Chapter. 2]{russell2020artificial}.
It is important to note that an episode is a sequential set of states from the agent in imitation learning literature, while \citeauthor{russell2020artificial} classify an episode as a single step from the agent (an experience).
In episodic environments, the task divides the agent's experience into atomic experiences.
The agent receives a state in each experience and then performs a single action. 
Additionally, the following experience is independent of the actions taken in the previous one.
However, this characteristic depends on how the environment computes these experiences.
For example, an experience usually comprises four frames in Atari games, while it is a single frame in Mario games.
Therefore, being episodic has some space for interpretation. 
 
Precision environments usually involve robotics tasks where agents require precision, and time for each decision is irrelevant.
The most common precision environment type is robotic arms.
This simulation works for various tasks, such as opening doors~\cite{torabi2021dealio}, pushing pins~\cite{calinon2007what}, and arranging blocks~\cite{yan2017oneshot}. 
The most common setting for these tasks is a robotic arm with an unmovable base and some range of movement on other joints (Fig.~\ref{fig:sub:pandas}).
In these tasks, researchers aim for agents to overimitate a teacher in the same setting and adapt slightly in newer settings, such as posing blocks in different positions~\cite{yan2017oneshot} or moving with different body configurations~\cite{peng2018variational}.

Robot walking and navigation is also common tasks for precision environments.
For this task, the environment simulates a robot's body, such as MuJoCo, but to a higher degree of detail. 
The idea is to test whether the agent can navigate different obstacles and spaces.
Furthermore, the agent can have different body configurations, from vacuums to bipedal designs.
For example, micro air vehicles, or MAVs, is a common domain for these precision environments (Fig.~\ref{fig:sub:mav}), where the agent has to fly through obstacles without getting hit.
Some flying environments consider different seasons to understand how agents adapt to variations in images~\cite{daftry2017mav}.

Usually, work that use precision environments experiment with physical robots as the final evaluation step~\cite{rodrigo2023stable,paolillo2023dynamical,yang2023watch,pathak2018zeroshot,liu2018raw,zhang2018deep,daftry2017mav,finn2016guided}.
However, this approach is expensive, given the requirement of a robotic arm or different robot settings.
Therefore, some authors~\cite{pavse2020ridm} rely on simulations to test different settings, such as blocks world~\cite{yan2017oneshot}, where the robotic arm has to arrange blocks in order, and the authors split some initial settings and goals only for evaluation, making sure the agent is not only overimitating and has learned the underlying task correctly.
Table~\ref{tab:env_precision} shows all precision environments with some information regarding the environments'~attributes.

\subsection{Sequential environments}

Sequential environments could have long-lasting consequences, given the agent's prior actions.
They are the opposite of precision environments.
However, it is essential to observe that not all actions must impact future experiences.
These environments are harder to solve for imitation learning agents because they require the agent to plan for future actions, which might be learning an abstraction from the task, and usually rely on non-stationary policies. 
For example, in Lunar Lander, an environment where a rocket has to use its engines to land between two flags, the agent have to learn how to map a trajectory depending on where the agent spawns.
If we suppose the agent overimitates teachers' trajectories instead of adapting actions to consider the difference between previous demonstrations and the current setting, the agent will not learn how to land the rocket where it is suppose to, but instead, it will learn how to land the rocket where the teacher landed it.

\begin{figure}[b]
    \centering
    \begin{subfigure}[b]{0.15\textwidth}
        \centering
        \includegraphics[width=\textwidth]{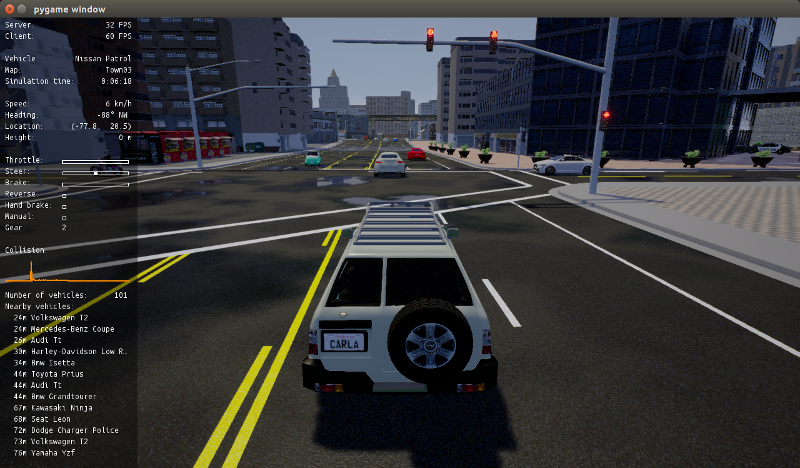}
        \caption{CARLA~\cite{dosovitskiy17carla}}
        \label{fig:sub:carla}
    \end{subfigure}
    \hfill
    \begin{subfigure}[b]{0.15\textwidth}
        \centering
        \includegraphics[width=\textwidth]{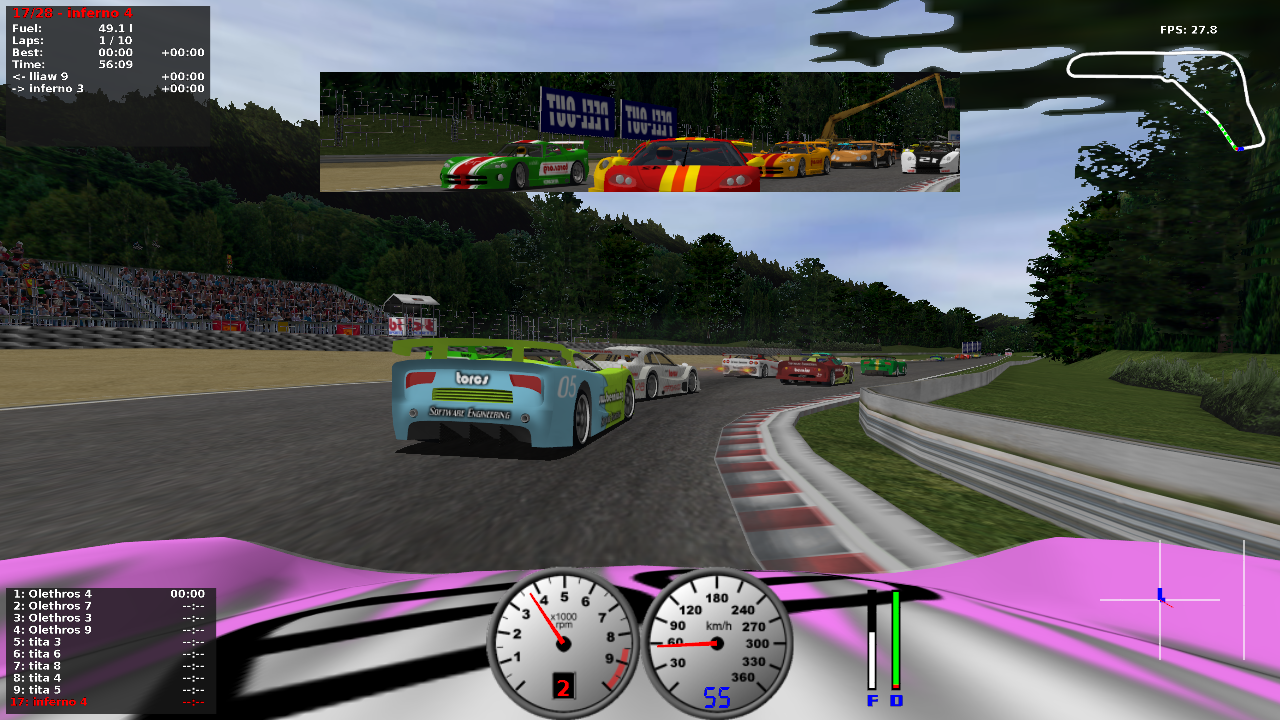}
        \caption{TORC~\cite{wymann2014torcs}}
        \label{fig:sub:torc}
    \end{subfigure}
    \hfill
    \begin{subfigure}[b]{0.17\textwidth}
        \centering
        \includegraphics[width=\textwidth]{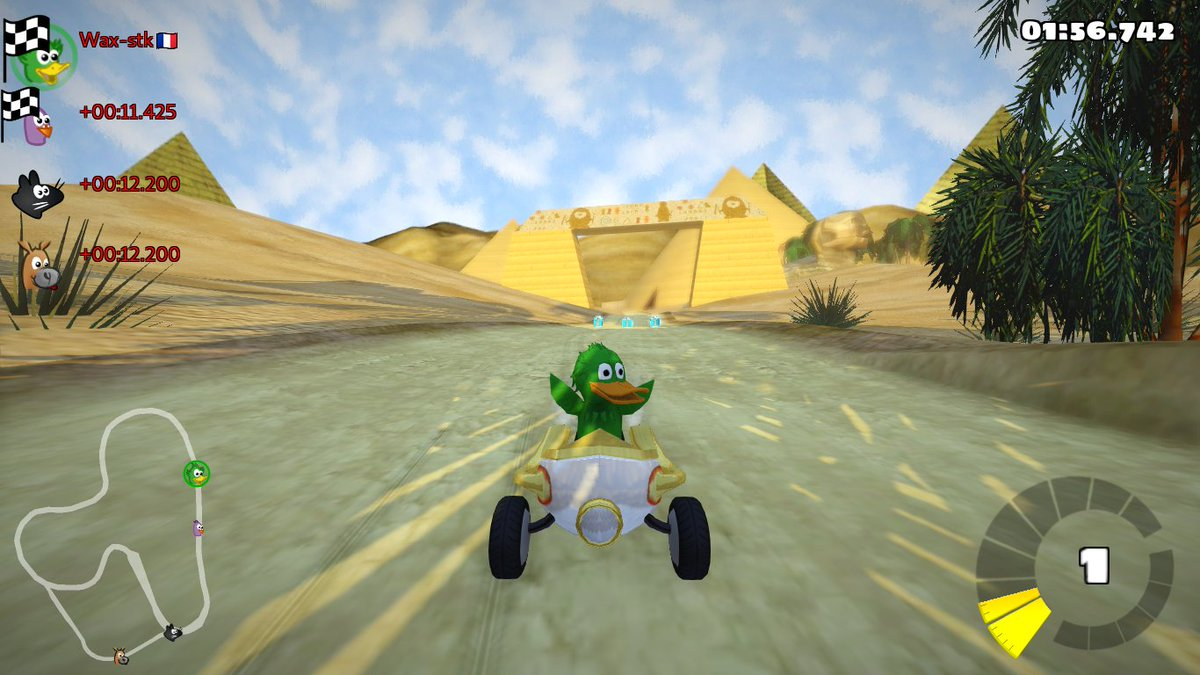}
        \caption{SuperTuxKart~\cite{henrichs2006stk}}
        \label{fig:sub:stk}
    \end{subfigure}
    \hfill
    \begin{subfigure}[b]{0.15\textwidth}
        \centering
        \includegraphics[width=\textwidth]{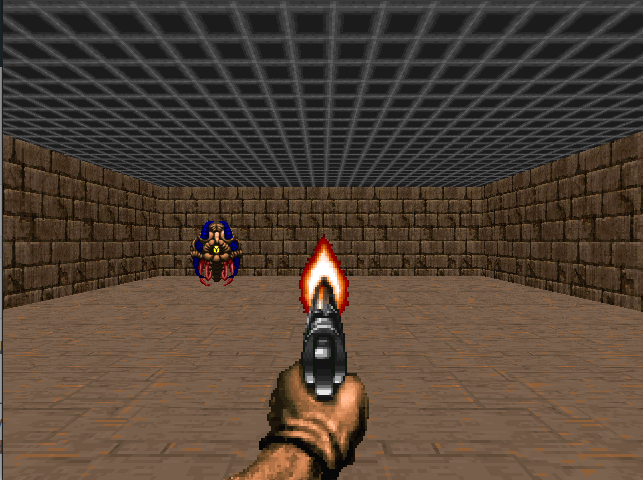}
        \caption{DOOM~\cite{Kempka2016ViZDoom}}
        \label{fig:sub:doom}
    \end{subfigure}
    \hfill
    \begin{subfigure}[b]{0.15\textwidth}
        \centering
        \includegraphics[width=\textwidth]{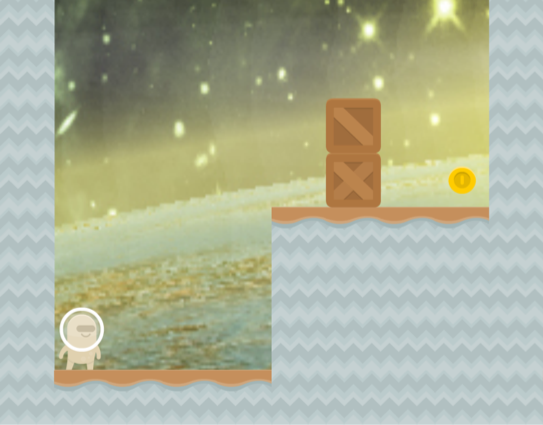}
        \caption{CoinRun~\cite{cobbe2019coinrun}}
        \label{fig:sub:coinrun}
    \end{subfigure}
    \hfill
    \begin{subfigure}[b]{0.15\textwidth}
        \centering
        \includegraphics[width=\textwidth]{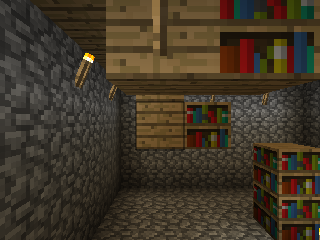}
        \caption{Minecraft~\cite{fan2022minedojo}}
        \label{fig:sub:minecraft}
    \end{subfigure}
    \caption{Examples of sequential environments.}
    \Description{Examples of sequential environments.}
    \label{fig:sequential}
\end{figure}

Sequential environments are most commonly used for autonomous driving domains, where the agent must learn how to drive and navigate different scenarios.
In these environments, different aspects can be experimented with.
For example, while CARLA~\cite{dosovitskiy17carla} has a more realistic city traffic simulation, TORCS~\cite{wymann2014torcs} proposes a scenario for open racing car simulation, where the agent competes against other cars (Fig.~\ref{fig:sub:carla} and~\ref{fig:sub:torc}, respectively).
Simulation of Urban Mobility, or SUMO~\cite{lopez2018sumo}, allows for recreating various urban scenarios, but it does not have CARLA's realistic LIDAR and RADAR sensors.
In contrast, SuperTuxKart offers a driving simulation with a more cartoonish style where the agent has different powers to use against other computer-controlled agents (Fig.~\ref{fig:sub:stk}).
Therefore, each environment can have drastic differences in information and settings, and researchers must ponder which scenario is more relevant to their research and correctly evaluates their agents. 

On the other hand, if the domain of autonomous driving is not required, the usual sequential environment is video games.
Video games, such as Mario~\cite{kauten2018mario} and Overcooked~\cite{carroll2019overcooked}, allow for easy abstraction of tasks, such as navigation, or just for modelling temporal consequences.
In the latter, these environments allow for the manipulation of maps to test specific behaviours from the agent.
For example, VizDoom~\cite{Kempka2016ViZDoom} (Fig.~\ref{fig:sub:doom}) allows developing agents to play DOOM, a $1993$ video game where you play a marine fighting through hordes of monsters in first person, using visual information.
The environment has visual editors that allow for custom scenarios and to control variables for each level, allowing for a more consistent benchmark between researchers.
Another advantage of these environments is that they provide diverse contexts for both learning and assessment, such as changing the number of enemies, weapons and level design between training and evaluation.
CoinRun~\cite{cobbe2019coinrun} (Fig.~\ref{fig:sub:coinrun}), where the agent learns how to navigate through a level from a game where the goal is to pick one or more coins, poses a similar setting.
During training, the level has no hindrances to the agent, but during the evaluation, the level will have monsters and lava, which can kill the agent.
Consequently, sequential environments are great for evaluating not only long-lasting consequences but also the agent's ability to generalise to different~settings.

Nevertheless, this diversity of settings for sequential environments can cause problems when comparing different work.
For example, in \citeauthor{yan2017oneshot}'s work use a maze to evaluate whether their agent learned the underlying task~\cite{yan2017oneshot}.
However, in \citeauthor{gavenski2020iupe}'s the authors use a different maze to evaluate their agent~\cite{gavenski2020iupe}.
In the first, the authors use a more simplistic structure (following a `C' and `S' shape).
The latter, uses a more complex structure, with more turns and dead ends with different sizes ($3 \times 3$, $5 \times 5$, and $10 \times 10$).
Thus, making the comparison between these two work harder.

Combining sequential and validation environments is ideal when creating imitation learning agents.
These long-lasting consequences usually are the downfall of these agents.
More so, these environments allow for diverse customisation of different levels of requirements.
Table~\ref{tab:env_sequential} shows all sequential environments with some information regarding the environments' attributes.

\subsection{Environments Discussion}

\begin{figure}[b]
    \centering
    \begin{minipage}[b][][b]{0.49\textwidth}
        \centering
        \includegraphics[width=\linewidth]{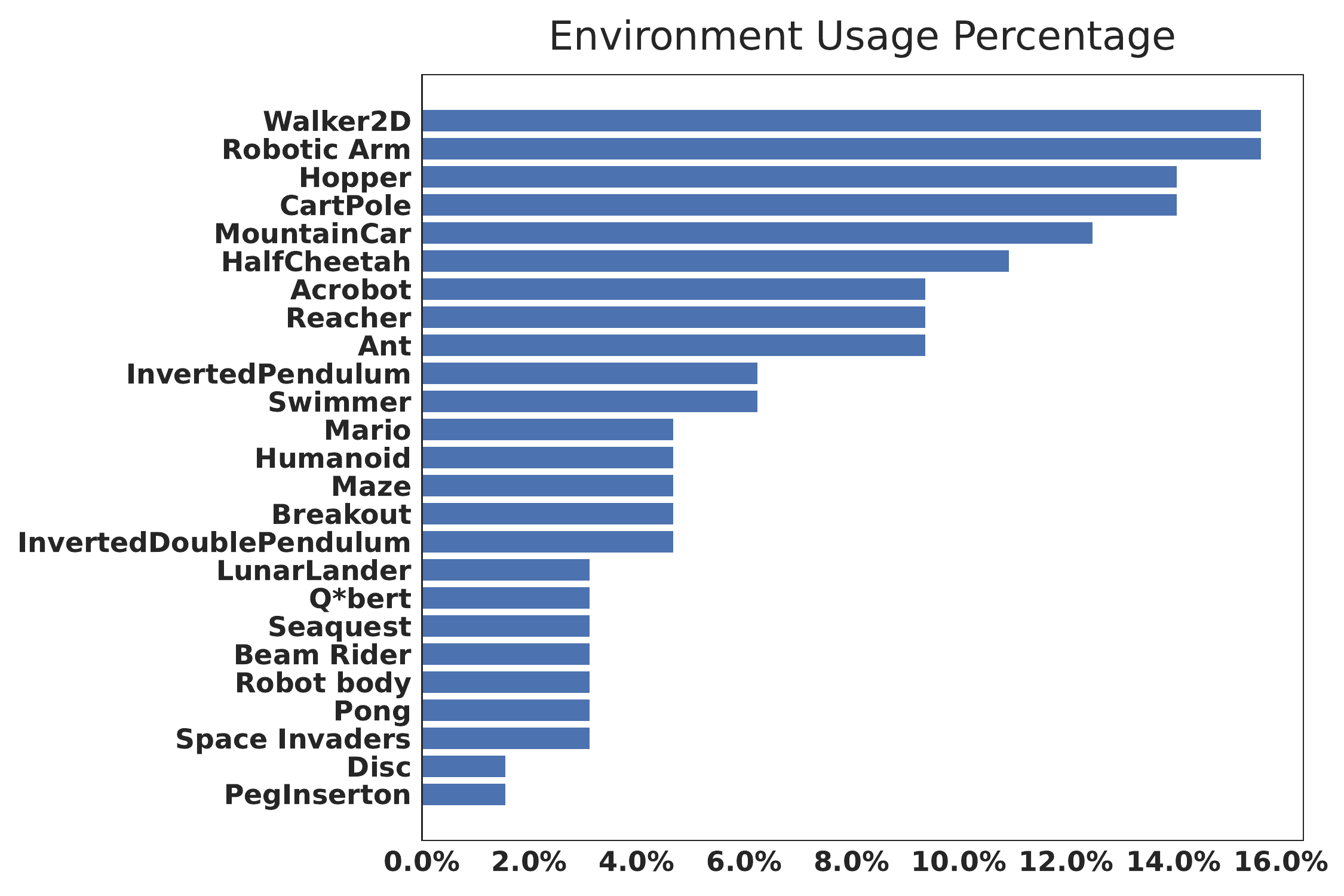}
        \caption{$25$ most used environments distribution.}
        \Description{25 most used environments distribution.}
        \label{fig:environments:usage}
    \end{minipage}
    \hfill
    \nextfloat
    \begin{minipage}[b][][c]{0.49\textwidth}
        \centering
        \begin{tikzpicture}[
    node distance=3cm,
    every node/.style = {
        rectangle, 
        rounded corners,
        minimum height=1cm,
        font=\Large \bfseries,
        text centered
    },
    node/.style = {
        minimum width = 5cm
    },
    subnode/.style = {
        minimum width = 4cm
    },
    arrow/.style = {
        ultra thick,
        ->,
        >=latex
    }
]

    \node (root) [
        node, fill=white!10, draw=black, minimum width={width("Imitation Learning Metrics (50)") + 100pt}
    ] {Imitation Learning Metrics (50)};

    \node (domain) [node, below of=root, fill=cb_blue!25, draw=cb_blue!75] {Domain (15)};
    \node (dom_quantitative) [%
        subnode, 
        below of=domain, 
        fill=cb_blue!25, 
        draw=cb_blue!90, 
        node distance=1.5cm, 
        xshift=1cm,
        align=center
    ] {Quantitative (14)};
    \node (dom_qualitative) [%
        subnode, 
        below=of dom_quantitative.west, 
        fill=cb_blue!25, 
        draw=cb_blue!75, 
        yshift=1.5cm,
        anchor=west,
        align=center
    ] {Qualitative (1)};

    \node (behaviour) [node, left of=domain, xshift=-3cm, fill=cb_orange!10, draw=cb_orange!50] {Behaviour (25)};
    \node (beh_quantitative) [%
        subnode, 
        below of=behaviour, 
        fill=cb_orange!10, 
        draw=cb_orange!50, 
        node distance=1.5cm, 
        xshift=1cm,
        align=center
    ] {Quantitative (18)};
    \node (beh_qualitative) [%
        subnode, 
        below=of beh_quantitative.west,
        fill=cb_orange!10,
        draw=cb_orange!50,
        yshift=1.5cm,
        anchor=west,
        align=center
    ] {Qualitative (7)};

    \node (model) [node, right of=domain, xshift=3cm, fill=cb_purple!10, draw=cb_purple!50] {Model (10)};
    \node (mod_quantitative) [%
        subnode, 
        below of=model, 
        fill=cb_purple!10, 
        draw=cb_purple!50, 
        node distance=1.5cm, 
        xshift=1cm,
        align=center
    ] {Quantitative (8)};
    \node (mod_qualitative) [%
        subnode, 
        below=of mod_quantitative.west,
        fill=cb_purple!10,
        draw=cb_purple!50,
        yshift=1.5cm,
        anchor=west,
        align=center
    ] {Qualitative (2)};

    \draw [arrow]   (root) -- (domain);
    \draw [arrow]   ++(0, -1.25cm) -|
                    (behaviour.north);
    \draw [arrow]   ++(0, -1.25cm) -|
                    (model.north);

    \draw [arrow]   ([xshift=-1.5cm]domain.south) |- (dom_quantitative.west);
    \draw [arrow]   ([xshift=-1.5cm]domain.south) |- (dom_qualitative.west);

    \draw [arrow]   ([xshift=-1.5cm]behaviour.south) |- (beh_quantitative.west);
    \draw [arrow]   ([xshift=-1.5cm]behaviour.south) |- (beh_qualitative.west);

    \draw [arrow]   ([xshift=-1.5cm]model.south) |- (mod_quantitative.west);
    \draw [arrow]   ([xshift=-1.5cm]model.south) |- (mod_qualitative.west);

\end{tikzpicture}
        \vspace*{0.6cm}
        \caption{Imitation learning metrics taxonomy.}
        \Description{Imitation learning metrics taxonomy.}
        \label{fig:metrics:taxonomy}
    \end{minipage}
\end{figure}

In this section we analysed all environments used in the literature and classified them in three different categories: validation, precision, and sequential.
In their usage, we observe a behaviour similar to Zipf's law, where from the $66$ environments used to evaluate the various imitation learning work (presented in Tables \ref{tab:env_validation}, \ref{tab:env_precision}, and \ref{tab:env_sequential}), six environments account for more than $80\%$ of the total number of times these environments were used.
Fig.~\ref{fig:environments:usage} illustrates this behaviour by displaying the distribution of the environments used in the literature.
Walker-2D, MuJoCo robotic arm simulation, Hopper, CartPole, MountainCar and HalfCheetah are the most common environments.
We classify five of these six as validation environments, except for the robotic arm.
As pointed out in Section~\ref{sub:validation}, this trend worries us since using only validation environments might incur the wrong evaluation of the imitation learning agent.
Nevertheless, $42$ out of the $66$ environments appear only once in the literature.
This behaviour is worrisome because it shows a lack of experimental protocols. 
We recognise that various tasks and domains may require particular settings. 
However, when analysing the work for autonomous vehicles, for example, we observe that the simulations were used only once on all domain-relevant work.
Furthermore, \citeauthor{zheng2021imitation} also point out this behaviour in their survey~\cite{zheng2021imitation}, when presenting the high-level and low-level tasks, where they argue that researchers have sometimes might bias their evaluation by selecting different environments.
We attribute this behaviour to reviewers and researchers looking to evaluation metrics and not to evaluation protocols, which shows a need for a more formal definition of more thorough protocols.

\section{Imitation Learning Metrics} \label{sec:metrics}

Imitation learning evaluation metrics can drastically differ from each work.
Some work rely on more domain-specific metrics, while others present metrics that measure how the agent performs compared to its teacher.
The most common taxonomy for evaluation metrics is the quantitative and qualitative classification.
However, this differentiation fails to consider the context of a metric.
For example, a metric that evaluates the distance between an agent's trajectory and its teacher in a task requiring precision will convey that the agent is performing well.
Conversely, if the environment requires less precision and a goal state holds higher importance, using a metric that measures an entire trajectory might convey that an agent that does not reach its goal state but follows the teacher's trajectory until a point, such as stopping right before the goal, is better than one that takes a significantly different trajectory but reaches the goal.
Therefore, we classify metrics as quantitative and qualitative but under behaviour, domain and model metrics.
Behaviour metrics are those that measure the agent's behaviour, they convey how distant an agent is from its teacher.
One might argue whether behaviour metrics measure teachers behaviour~\cite{gavenski2022how}, but we shy away from this discussion for now.
Domain metrics measure domain-specific properties; they bring a more contextual measurement for tasks with more specific requirements, such as how many traffic infractions an agent commits in a self-driving car environment.
Finally, model metrics measure the learning procedure of agents, such as how accurate a model is when predicting an action.

\subsection{Behaviour metrics} \label{ch:sec:sub:behaviour_metrics}

Behaviour metrics are the most common metrics among all imitation learning work.
Usually, they depend on the environment's reward or the distance from one property of the teacher.
The most used metrics are reward-based ones since comparing the reward from an agent to its teacher intuitively shows how the agents perform comparatively.
Three variations from the reward-based metrics are present in the reviewed literature: 
\begin{enumerate*}[label=(\roman*)]
    \item accumulated reward;
    \item average episodic reward; and
    \item performance.
\end{enumerate*}

\textit{Accumulated reward} is how much reward the agent accumulates in an episode.
We reiterate here that an episode in this work differentiates from \citeauthor{norvig2002modern}'s definition~\cite{norvig2002modern} and refers to a set of experiences from an agent in an environment.
The final value is given by summing the reward retrieved from the Markov decision process in an episode with $n$ steps (Equation~\ref{eq:acc_reward}).
This metric conveys how the agent performs the environment task and allows researchers to compare student and teacher rewards.
However, using a single episode might be misleading.
For example, suppose the \textit{seed} (a random numerical value used to initialise the environment) used in the evaluation is present in the teacher dataset. 
In that case, the agent must only match the given states to those in the training set to achieve the same reward.
Moreover, even when the seed is absent in the teacher dataset, some initial states might be closer to the teacher's initial state than others.
Thus, some states require less generalisation from the agent than others, making some initial states `easier' than others.
Therefore, the accumulated reward might not be the best metric to measure how well the agent generalises in some unfavourable instances.

\begin{equation} \label{eq:acc_reward}
    AR(\policy) = \sum_{i=1}^{n} \gamma^{n} r(s_i, \policy(s_i))
\end{equation}

A more statistical approach would be to use the average reward for a set number of episodes and study the deviation between each experience.
Therefore, the chance of finding diverse seeds increases, and the standard deviation among all episodes can show whether the agent is consistent among different experiences.
The average of all accumulated rewards is called the \textit{average episodic reward}.
In it, the agent records its accumulated reward for $k$ number of episodes, and the average is calculated (Equation~\ref{eq:aer}). 
The number of episodes is usually $100$ or $10$, and the standard deviation is shown alongside it.
The average episodic reward has the same benefits as the accumulated reward and solves the problem from untested generalisation, as long as there are enough episodes.
However, it can be the case that an agent's average episodic reward is closer to that of a random initialised agent than to that of a teacher in a specific task.
To understand how an agent compares between all agents, all three reward values would have to be given.

\begin{equation} \label{eq:aer}
    AER(\policy) = \frac{1}{k}\sum^{k}_{j=1} AR(\policy)
\end{equation}

\textit{Performance} solves the comparison issue by applying normalisation in the accumulated reward.
It uses a random policy's $\random$ accumulated reward as the minimum and the teacher's as the maximum (Equation~\ref{eq:perf}).
In this metric, if an agent accumulates as much reward as a random one, the performance will be $0$.
However, if the same agent accumulates the same reward as the teacher, its performance will be $1$.
An agent can achieve a performance higher than $1$ when it accumulates more than its teacher and lower than $0$ when its rewards are inferior to a random agent.
The performance metric is usually retrieved from $k$ trajectories to get the same benefits the average episodic reward has.
However, some work directly uses the average episodic reward instead of the accumulated reward when computing the metric (Equation~\ref{eq:perf_aer}).

\begin{figure}[h]
    \centering
    \begin{minipage}{.49\linewidth}
        \begin{equation} \label{eq:perf}
            \mathcal{P}_\tau(\policy) = \frac{AR(\policy) - AR(\random)}{AR(\teacher) - AR(\random)}
        \end{equation}
    \end{minipage}%
    \hfill
    \begin{minipage}{.49\linewidth}
        \begin{equation} \label{eq:perf_aer}
            \mathcal{P}_\Tau(\policy) = \frac{AER(\policy) - AER(\random)}{AER(\teacher) - AER(\random)}
        \end{equation}
    \end{minipage}%
    \Description{Equations.}
\end{figure}

Even though the performance metric has all the benefits of the other reward-based metrics, it inherits the same issues from linear normalisation procedures when dealing with skewed data~\cite{gavenski2022how}.
For example, suppose we assume that the minimum reward (random) is $0$, the maximum (teacher) is $100$. The mean score from agents is $50$, and the median score is $20$.
In this case, the majority of the performance for the agents will be clustered towards the minimum, and there will be very little differentiation among them.
Thus, performance becomes a less useful metric when the agent's performance is close.

Although the metric does not optimise the agent, it can misrepresent the agent's behaviour in a given environment.
For example, imagine two scenarios for an environment where the threshold to be considered optimal is a $100$ reward:
\begin{enumerate*}[label=(\roman*)]
    \item the teacher reward is $75$, and the random reward is $0$; and
    \item the teacher reward is $100$, and the random reward is again $0$.
\end{enumerate*}
In the first scenario, when an agent achieves teacher behaviour, it will score a performance of $1$, even though the agent was $25$ points short from an optimal result.
By achieving the same reward from the first scenario ($75$) in the second scenario, the agent will only achieve a performance of $0.75$.
This example shows the problem of assuming as a premise that the teacher is an optimal policy.
If the selected teacher is not optimal, the performance metric will not accurately represent how good the agent is.
Moreover, in the first scenario, although the agent is $25$ points from being considered optimal for the task, the metric conveys optimality.
Hence, there is a need to know beforehand how optimal the teacher is, which can be a complex problem in several application domains.
On the other hand, if an agent achieves a reward of $150$ in the same example, the performance will be $2$ and $1.5$ for the first and second scenarios, respectively.
Although the agent achieves performance higher than the teacher, this result shows that:
\begin{enumerate*}[label=(\roman*)]
    \item the samples might not be ideal; and 
    \item the policy might not have learned the exact behaviour from the teacher.
\end{enumerate*}

Distance metrics are the most common metrics after reward-based ones.
They usually involve measuring how far the agent is from one or more properties of the teacher.
For example, some work measures the distance between the teacher's and agent's trajectories or between the artificial reward function and the MDP's $r$. 
Although the distance metric can vary, the most common distances are the \textit{Manhattan distance} (Equation~\ref{eq:minkowski}, where $p=1$) and \textit{Euclidean distance} (Equation~\ref{eq:minkowski}, where $p=2$), where an input $I$ may be single states, trajectories or state-action pairs, and the \textit{Kullback-Leibler divergence} (Equation~\ref{eq:kl}), where $P$ and $Q$ are two probability functions.
A common issue with distance-based metrics is the requirement for teachers to be accessible to measure the chosen property (when observations do not provide the desired property). 
These metrics require some higher degree of control for measuring the same property for the agent. 
An example would be measuring the precision of unseen trajectories.
One might reserve part of the trajectory data to measure how close agents and teachers are.
However, this setup would require a more significant number of data from researchers.
A second issue with distance-based metrics is that they might misrepresent the behaviour of an agent due to compounding errors from different actions during the early stages of an episode.
Suppose an agent acts differently from its teacher earlier on. 
In that case, the action might cause the trajectory to be further apart even though the agent's behaviour might be close to the teacher after these earlier steps.
This approach to measuring would require researchers to have a tree of all possible teacher's actions to compare the agent's actions given a diverging path from its initial data.

\begin{figure}[h]
    \centering
    \begin{minipage}{.49\linewidth}  
        \begin{equation}\label{eq:minkowski}
            d(I_\agent, I_\teacher, p) = \left( \sum_{i=1}^{n} \left| I^i_\agent - I^i_\teacher \right|^p \right)^\frac{1}{p}    
        \end{equation}
    \end{minipage}
    \hfill
    \begin{minipage}{.49\linewidth}
        \begin{equation} \label{eq:kl}
            d(P \mid \mid Q) = \sum_{s \in S} P(s) \ln \left ( \frac{P(s)}{Q(s)} \right )
        \end{equation}
    \end{minipage}
    \Description{Equations.}
\end{figure}

The \textit{success rate} ($SR$) also appears as a viable quantitative behavioural metric.
In it, the authors usually select a goal $g$ for the environment (it could be the environment's goal or a user-defined one) and collect a set of trajectories $\Tau$, one for each episode, and check whether each trajectory $\tau$ has $g$ in it (Equation~\ref{eq:sr}).

\begin{figure}[h]
    \centering
    \begin{minipage}{.49\linewidth}
        \begin{equation} \label{eq:sr}
            SR(g, \Tau) = \frac{1}{\mid \Tau \mid} \sum_{\tau \in \Tau} G(g, \tau), \text{ where}
        \end{equation}
    \end{minipage}
    \hfill
    \begin{minipage}{.49\linewidth}
        \begin{equation}
            G(g, \tau) = \left\{\begin{matrix}
                1 & \forall_n s_n \in \tau \mid \exists_t s_n = g\\ 
                0 & otherwise
        \end{matrix}\right.
        \end{equation}
    \end{minipage}
    \Description{Equations.}
\end{figure}

\noindent
Coupling a reward-based and distance-based metric with a success rate can help researchers understand how the agent behaves.
Both initial metrics convey information about the agent's trajectory while the latter conveys if the agent achieves its intended goal.

Nevertheless, these metrics do not have information regarding how human-like the agent behaviour is (considering the agent trained with human demonstrations).
Therefore, some work employ qualitative metrics like questionnaires and Turing tests.
For questionnaires, questions are usually regarding how similar to humans the agents look~\cite{lee2014mario} or how easy it is to play with the agent~\cite{xingzhou2023pecan}.
In the Turing Test, participants classify if a video from a agent interacting with the environment is human or agent.
Thus, the agent aims to deceive the human participants (similar to adversarial learning).
Moreover, other researchers~\cite{daftry2017mav,finn2016guided,calinon2007what} also observe the agent's trajectory and do some qualitative analysis.
More related to robotics, this analysis usually displays a graph with various trajectories for the robot's movement, and the authors explain why the divergence in trajectory is good or acceptable in a precise task.
Unfortunately, the qualitative metrics that require human participants are costly in comparison to the quantitative ones.
Thus, they are less common in more recent work.
Although these metrics are essential for evaluating imitation learning agents, we understand that to use qualitative metrics requires significant effort. 
Creating a questionnaire requires question design, which is uncommon, and the Turing test requires participants to watch numerous videos of humans and agents playing to vote.

\subsection{Domain metrics}
It is not always the case that behaviour metrics convey the information required to evaluate whether an agent performs adequately in an environment.
Some domains allow the usage of specific metrics to measure the optimality of the agent.
These metrics are specific to these domains and do not make sense to use anywhere else.
The most common domain-specific metrics are those used in autonomous driving and robotics.

For driving tasks, different behaviours from the agent can be measured, such as distance travelled~\cite{ross2011reduction,daftry2017mav}, traffic infractions~\cite{chen2020cheating}, time to collision and distance headway~\cite{hu2023safe}, allowing for more accurate measurement of how safely the agent navigates through the environment.
Even though the reward function could contain this information, by separating them, the researcher can better comprehend the agent's behaviour and try to optimise different things without altering the reward function to represent the desired behaviour better.

In robotics, most domain-specific metrics measure time~\cite{dambrosio2013scalable,daftry2017mav,ross2011reduction} or failure rate~\cite{ross2011reduction,nair2017combining}.
In the case of time, these metrics convey how efficient the robot arm is in performing a task.
For example, how long it takes to order a set of blocks, or how precise the robot can perform a variety of movements, such as the pixel difference between teacher and agent in the same settings.
For failure rate, most metrics display the percentage of times the robot failed to reach its goal, such as running in a simulation.
Conversely, we did not find any work that gave intuition behind the failure, which is crucial to understanding whether the final result is good or bad.
We understand that doing this analysis is more costly and adds a qualitative tone to it, which the domain-specific metrics lack; however, only providing a number representing failure, such as how many times the robot fell during a run, makes it hard to compare.
Compared to humans, competitors almost never fall during a running competition, and to other robots, the results might be significant.

Considering that imitation learning agents do not use reward functions to learn, it would not be very sensible to require a reward function to measure how optimal an agent is.
Therefore, some work use domain-specific scoring to show how an agent performs.
For example, in some environments, researchers use how many coins~\cite{edwards2019ilpo,lee2014mario} the agent collects during an episode or how many goals a team scores~\cite{raza2012il,dambrosio2013scalable}.
Nevertheless, these metrics lack the finesse of other metrics and should be used with behavioural- or model-specific metrics.

\subsection{Model metrics}

Model metrics are the usual metrics we see in learning research.
They convey how accurate or inaccurate a model is.
Most work use \textit{accuracy} since imitation learning approaches use some form of behavioural cloning, which is a supervised-learning approach, where accuracy is defined as the ratio between the number of correct predictions and the total number of predictions.
Other work use the \textit{error rate} from their models, which usually rely on generative models, such as forward dynamics models.
However, these metrics do not convey useful information regarding the agent's performance in a task.
An example is the accuracy of inverse dynamics models.
Although they can have high accuracy in their demonstration datasets, they might have poor accuracy in the teacher's observation since it usually follows other distributions.
Therefore, accuracy might give the wrong impression that the model is classifying each transition correctly when it is not.
Model-specific metrics are used during training to understand whether a model can learn the desired task.
Additionally, some work also uses these metrics to interpret when a model becomes biased toward a set of actions~\cite{gavenski2020iupe}.


On the other hand, some work~\cite{yusuf2018playing,yunzhu2017infogail} use other model metrics to have intuition behind the model's learning.
In these work, the authors use dimensionality reduction techniques, such as \textit{t-SNE}~\cite{maaten2008tsne}, to interpret how good their models are in clustering different actions.
Although these techniques are promising to give some qualitative intuition behind the models' performance in clustering information, these metrics are often used when the work focuses outside imitation learning, for example, transfer learning. 
Moreover, when using these dimensionality reduction techniques, one should consider that: 
\begin{enumerate*}[label=(\roman*)]
    \item they are highly sensitive to hyperparameters, as pointed out by \citeauthor{maaten2008tsne}~\cite{maaten2008tsne};
    \item interpreting them might be challenging~\cite{wattenberg2016how}, which makes them less suitable as an explainable method; and 
    \item and some of these techniques lack robustness and might not provide consistent results for similar datasets~\cite{becht2018evaluation}.
\end{enumerate*}

\begin{figure}[h]
    \centering
    \hfill
    \begin{minipage}{.4\textwidth}
        \centering
        \footnotesize
        \begin{algorithm}[H]
        \caption{Default evaluation protocol}
        \label{alg:evaluation}
        \begin{algorithmic}[1]
            \Require $D \gets \left\{(s_0, a_0);\cdots;(s_{n}, a_{n})\right\}$ \label{alg:line:teacher}
            \State train $\pi_\theta$ with $\Tau$ \label{alg:line:train}
            \State $AER_\agent \gets 0$
            \State $\mathcal{P}_\agent \gets 0$
            \For {$n \gets 1$ to $N$} \label{alg:line:loop}
                \State $AER_\agent \gets AER_\agent + \nicefrac{AR(\agent)}{N}$ \label{alg:line:aer}
                \State $\mathcal{P}_\agent \gets \mathcal{P}_\agent + \nicefrac{\mathcal{P}(\agent)}{N}$ \label{alg:line:p}
            \EndFor
        \end{algorithmic}
    \end{algorithm}
    \end{minipage}
    \hfill
    \begin{minipage}{0.5\textwidth}%
        \centering
        \begin{subfigure}[b]{0.3\textwidth}
            \centering
            \includegraphics[width=.9\linewidth]{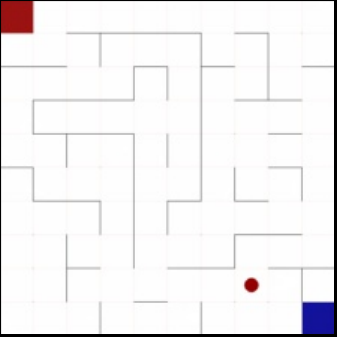}
            \caption{Input image.}
            \label{fig:sub:maze}
        \end{subfigure}%
        \hspace{10pt}%
        \begin{subfigure}[b]{0.3\textwidth}
            \centering
            \includegraphics[width=.9\linewidth]{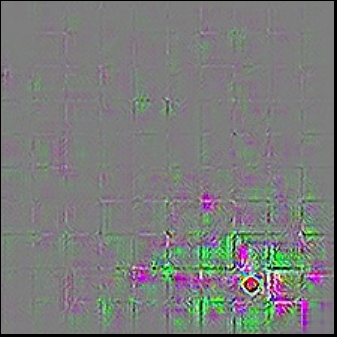}
            \caption{Saliency map.}
            \label{fig:sub:saliency}
        \end{subfigure}%
        \nextfloat
        \caption{Saliency map example from \citeauthor{gavenski2020iupe}'s work~\cite{gavenski2020iupe}.}
        \Description{Saliency map example from \citeauthor{gavenski2020iupe}'s work~\cite{gavenski2020iupe}.}
        \label{fig:saliency}        
    \end{minipage}
    \hfill
\end{figure}

\textit{Saliency maps} are also used as qualitative metrics in imitation learning literature~\cite{gavenski2020iupe,peng2018variational} as model-domain metrics.
They display in the model's input where does are the most active gradients.
It gives some intuition on what the model focuses on when predicting an action.
Fig.~\ref{fig:saliency} displays an example where the authors use the input image (Fig.~\ref{fig:sub:maze}) to display the attention map (Fig.~\ref{fig:sub:saliency}) to understand what the agent may be using to classify the action.
On the other hand, saliency maps share three of the major drawbacks of dimensionality reduction techniques, which we understand we need to focus on:
\begin{enumerate*}[label=(\roman*)]
    \item subjectivity;
    \item unreliability; and
    \item interpretability.
\end{enumerate*}
The first comes from the fact that saliency maps are influenced by the choice of the saliency model and parameters used in their generation.
Different models may produce varying results for the same input, and there is no consensus on which model is the best~\cite{bylinskii2019evaluation}.
Given the subjectivity of saliency maps, it can be sensible to assume that such behaviour might cause unreliable results~\cite{kindermans2019saliency,adebayo2018sanity}.
Although these models can produce impressive results when connecting labels with raw input data, connecting more complex knowledge representations might be difficult, especially when connecting internal knowledge representations from neural networks to humans~\cite{holmberg2022more}.
Finally, saliency maps might be challenging to interpret~\cite{montavon2018methods}.
When interpreting these results, one should remember that there are different possible conclusions given a combination of input and label.

Therefore, researchers should consider using these model metrics to provide an intuition of what imitation learning agents learned or how they behave, but not as concrete proof.
Furthermore, a combination of these metrics with domain-specific and behaviour-specific metrics, which show a more accurate representation of how optimal the agent is, should be present for a more complete conclusion.

\subsection{Evaluation protocol} \label{sec:sub:evaluation:protocol}

All work presented in Section~\ref{sec:methods} follow in some form the structure in Algorithm~\ref{alg:evaluation} when evaluating their method.
They train a policy $\agent$, given a set of teacher demonstrations $D$ (Line~\ref{alg:line:train}).
For simplicity, here we consider that all samples are demonstrations (have action notations) (Line~\ref{alg:line:teacher}), but one should assume that this is only true for learning from demonstration scenarios since learning from observation does not require action information during training.
Algorithm~\ref{alg:evaluation} also only considers trained policies since it refers to the final evaluation protocol, so we remove any iterative training notation.
Finally, researchers report using some metrics presented in this section.
The number of episodes from an agent in an environment ($N$) varies between $5$ to $100$.
Although imitation learning work rarely inform how the seeds are selected.
It is sensible to assume that the seeds used to collect teacher demonstrations are absent from the evaluation process.
Unfortunately, this assumption is not always true since the newest version from OpenAI Gym~\cite{zenodo2023gymnasium} removes the option for users to set a specific seed for the environment.
Therefore, some work use the same seed for both teacher and agent, which can lead to unfair~comparisons.

Moreover, some methods employ other experiments to evaluate their performance in scenarios with different amounts of samples (sample efficiency)~\cite{monteiro2023self,gavenski2021self,kidambi2021mobile,zhu2020opolo}.
In these scenarios, researchers reduce the number of teacher episodes the agent has access to and perform the same evaluation protocol in Algorithm~\ref{alg:evaluation}.
Other work experiment with different degrees of optimality and measure each method's robustness~\cite{gavenski2022how}.
In it, they create teacher samples with different degrees of optimality and perform the evaluation protocol in Algorithm~\ref{alg:evaluation}.

Finally, to compare their work, other approaches run the same steps from Algorithm~\ref{alg:evaluation} for all baselines selected.
Researchers frequently select the traditional behavioural cloning method as a benchmark alongside newer approaches in the same category as the proposed method.
For example, learning from observation methods usually compare their approach with behavioural cloning since it has access to the teacher's actions and works as a ground-truth comparison. 
Moreover, adversarial learning methods compare their approach with \citeauthor{ho2016generative}'s work~\cite{ho2016generative}, even though they are outdated.

\subsection{Metrics Discussion}

In this section, we listed different metrics with different contexts for imitation learning work.
We divided them into three different groups: behaviour, domain and model.
When studying these metrics, we found $50$ different ones (differentiating them by context and equation and not only by name).
As discussed in Section~\ref{ch:sec:sub:behaviour_metrics}, most metrics focused on quantitative analysis, with the most common one using reward to measure how optimal the agent is. 
As with the environments, the metrics also present many that appear once.
However, unlike environments, comparing all methods with a new metric presents fewer unfair comparison problems if researchers compare all methods using the same metric.
Additionally, we observe an increase in performance usage, which considers policies from random and teachers agents to measure optimality.


Qualitative metrics are rare in most novel work, which is understandable since the cost of creating questionnaires or running Turing tests might make some research unfeasible. 
However, the absence of qualitative metrics in the literature is problematic since it does not allow researchers to understand how human-like the agent is.
In this literature review, there are $109$ uses of quantitative metrics over $41$ different metrics.
In comparison, there are only $10$ uses of $9$ different qualitative metrics.

Finally, we analyze different protocols for evaluating imitation learning methods.
Most work follows the same guidelines when evaluating novel methods and baselines.
In its essence, the presented protocol offers a straightforward way to evaluate imitation learning methods.
However, researchers fail to show how their method compared outside of the optimal scenario by not applying the same protocol to measure other characteristics outside the overall performance, such as efficiency and effectiveness.
We conclude that following protocols presented in Section~\ref{sec:sub:evaluation:protocol}, such as our protocol where we reduce the degree of optimality of experts~\cite{gavenski2022how}, might alleviate some of the issues from the original protocol but will not solve them.
\section{Reflections} \label{sec:reflections}

In this section, we reflect on the insights we had while writing this survey.
These reflections can help researchers better understand imitation learning and its challenges, which we discuss further at the end of this section.
Importantly, we aim to help new researchers design imitation learning agents while avoiding common pitfalls.

For this section, we consider that under the loss function $\loss$, an agent learns how to maximise the return of $\loss$ to perform a task under certain conditions.
$\loss$ might be the accumulated value of $\mdp$'s reward function ($\sum_i \gamma^i r_{i+1}$) or capture some other specific desired behaviour.
For example, $\loss$ may want to minimise the number of vertices in the solution of a path-finding problem, or instead compute the overall length of the path according to the weight of the edges between vertices.
Training a policy $\policy$ with parameters $\theta$ and loss function $\loss$ yields the value function $v_\loss(\pi_\theta, s)$, which measures the quality of $\policy$ for each state $s$.
A policy $\policy$ with parameters $\theta_o$ is \textit{optimal} for $\loss$ when for all states $s$ and all parameters $\beta$, $v_\loss(\agent_o, s)~\geqslant~v_\loss(\pi_\beta, s)$.
Moreover, $\policy$ is \textit{theoretically optimal}, when for \textit{all} loss functions $h$ and all parameters
$\beta$, $v_\loss(\agent_o, s)~\geqslant~v_h(\pi_\beta, s)$.
In other words, even though an agent may optimal under a given function $\loss$, it may still under-perform under another loss function $h$.

\subsection{Imitation Learning Evaluation} \label{sec:problem}

When we consider IL systems, we are likely to compare them to machine learning (ML) and reinforcement learning (RL) approaches.
After all, IL methods are a relaxation of the RL approach into an ML supervised classification problem.
However, this comparison comes with some preconceptions that we want to address:
\begin{enumerate*}[label=(\roman*)]
    \item disparities over testing processes;
    \item divergence in benchmarking; and
    \item the main goal of IL approaches.
\end{enumerate*}

\begin{wrapfigure}{r}{.45\linewidth}
    \centering
    \begin{subfigure}[b]{0.2\columnwidth}
        \centering
        \begin{tikzpicture}[
    node distance=2cm,
    arrow/.style={
        ->,
        >=latex
    },
]
    \draw[step=1cm, color=black, opacity=0.3] (0, 0) grid (3, 3);

    \draw[draw=black,opacity=0.6] (0,0) rectangle ++(3,3);
    
    \draw [arrow, color=cb_purple] (0.7, 0.5) -- (1.5, 0.5) -- (2.5, 0.5) -- (2.5, 1.5) -- (2.5, 2.3)  node [pos=.72, xshift=-4pt,label=right:{\tiny $\nicefrac{1}{2}$}] {};
    \draw [arrow, color=cb_blue] (0.5, 0.7) -- (0.5, 1.5) -- (0.5, 2.5) -- (2.3, 2.5) node [pos=.92,yshift=-4pt, label=above:{\tiny $\nicefrac{1}{2}$}] {};

    \node at (0.5, 0.5) {$s_0$};
    \node at (1.5, 1.5) {$s_x$};
    \node at (0.5, 1.5) {$s_y$};
    \node at (1.5, 0.5) {$s_z$};
    \node at (2.5, 2.5) {$g$};
\end{tikzpicture}
        \caption{Optimal trajectories.}
        \label{fig:maze_all}
    \end{subfigure}
    \hspace{10pt}
    \begin{subfigure}[b]{0.2\columnwidth}
        \centering
        \begin{tikzpicture}[
    arrow/.style={
        ->,
        >=latex
    },
    bridging path/.initial=arc,
    bridging span/.initial=8pt,
    bridging gap/.initial=4pt,
    bridge/.style 2 args={
        spath/split at intersections with={#1}{#2},
        spath/insert gaps after
        components={#1}{\pgfkeysvalueof{/tikz/bridging span}},
        spath/join components upright
        with={#1}{\pgfkeysvalueof{/tikz/bridging path}},
        spath/split at intersections with={#2}{#1},
        spath/insert gaps after
        components={#2}{\pgfkeysvalueof{/tikz/bridging gap}},
    }
]
    \draw[step=1cm, color=black, opacity=0.3] (0, 0) grid (3, 3);

    \draw[draw=black,opacity=0.6] (0,0) rectangle ++(3,3);

    \node at (0.5, 0.5) {$s_0$};
    \node at (1.5, 1.5) {$s_x$};
    \node at (0.5, 1.5) {$s_y$};
    \node at (1.5, 0.5) {$s_z$};
    \node at (2.5, 2.5) {$g$};

    \path [spath/save=red_arrow] (0.3, 0.7) 
        -- (0.3, 1.7) 
        -- (1.5, 1.7) 
        -- (1.5, 2.4) 
        -- (2.3, 2.4) 
        node [pos=.95, yshift=-20pt, label=below:\textcolor{cb_blue}
        {\tiny $\nicefrac{3}{11}$}] {};
    
    \path [spath/save=blue_arrow] (0.7, 0.7) 
        -- (0.7, 1.5) 
        -- (2.4, 1.5) 
        -- (2.4, 2.3) 
        node [pos=.95, xshift=-6pt, label=left:\textcolor{cb_purple}
        {\tiny $\nicefrac{3}{11}$}] {};
        
    \path [spath/save=green_arrow] (0.7, 0.6) 
        -- (1.5, 0.6) 
        -- (1.5, 1.2) 
        -- (0.5, 1.2) 
        -- (0.5, 2.6) 
        -- (2.3, 2.6) 
        node [pos=.70, yshift=-6pt, label=above:\textcolor{cb_green}
        {\tiny $\nicefrac{4}{11}$}] {};
    
    \path [spath/save=purple_arrow] (0.7, 0.4)
        -- (2.6, 0.4)
        -- (2.6, 2.2) 
        node [pos=.2,xshift=-17pt, label=right:\textcolor{cb_orange}
        {\tiny $\nicefrac{1}{11}$}] {};

    \tikzset{bridge={green_arrow}{red_arrow}}
    \tikzset{bridge={blue_arrow}{green_arrow}}

    \draw [arrow, color=cb_purple, spath/use=red_arrow];   
    \draw [arrow, color=cb_blue, spath/use=blue_arrow]; 
    \draw [arrow, color=cb_green, spath/use=green_arrow]; 
    \draw [arrow, color=cb_orange, spath/use=purple_arrow]; 
\end{tikzpicture}
        \caption{Multiple trajectories.}
        \label{fig:maze_some}
    \end{subfigure}
    \caption{
        Examples for a teacher dataset with \textcolor{cb_blue}{blue}, \textcolor{cb_green}{green}, \textcolor{cb_orange}{orange} and \textcolor{cb_purple}{purple} trajectories.
        The numbers give the distribution of trajectories in the dataset. 
    }
    \label{fig:maze_fig}
\end{wrapfigure}
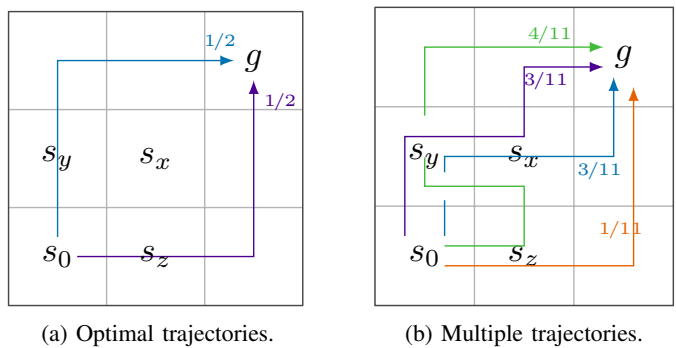

Traditional ML applications divide their data into three different group sets: training, validation and test sets.
They assume that all sets share the same distribution over data, and the test set should be only used at the end to evaluate how the model will perform under a `real-world' scenario.
This allows a well-defined and clear evaluation.
Conversely, IL approaches shy away from using test sets.
Since the typical application for IL approaches is agent-based, it is more practical to test the final learned policy via simulation of the environment.
This testing approach is similar to how RL methods are evaluated.
However, the training processes for RL and IL agents are different.
Training RL agents involves the use of exploration and exploitation phases without a training set.
During these phases, agents learn the different values for states outside an optimal trajectory.
Thus, even though a RL agent might not visit a state often, the agent will have some information about how to act on it.
The training of IL agents is limited to the states in the training set, which means they have no information about states outside this set.
For example, consider the grids in Fig.~\ref{fig:maze_fig}, where the objective is for the agent to reach square $g$.
Suppose the training set of an IL agent consists only of the two optimal trajectories in \textcolor{cb_purple}{purple} and \textcolor{cb_blue}{blue} given in Fig.~\ref{fig:maze_all}.
After training, the agent will have acquired some knowledge about how to get to $g$ from any square except for $s_x$. 
Furthermore, the probability of reaching $s_x$ is $0\%$ from the learned policy's perspective since the teacher samples prevent the agent from reaching $s_x$ from any state. 
Now consider the training set in Fig.~\ref{fig:maze_some}, containing  $4$ \textcolor{cb_green}{green}, $3$ \textcolor{cb_purple}{purple}, $3$ \textcolor{cb_blue}{blue}, and $1$ \textcolor{cb_orange}{orange} trajectories. The learned policy will not be successful for $50\%$ of the non-goal states, because the states $s_0$, $s_x$, $s_y$, and $s_z$ lead to an infinite loop. If the agent starts in $s_0$, it will go to $s_y$, since $P(\up\mid s_0) = 6/11$. 
If it starts in $s_z$, then it will go to $s_x$, since $P(\up\mid s_z)= 4/5$.
Once in $s_x$ or $s_y$, it will loop because from $s_x$, it will go to $s_y$, since $P(\lf\mid s_x) = 4/10$ (highest) and from $s_y$, it will go to $s_x$, since $P(\rg\mid s_y) = 6/10$. 
Therefore, it is important to understand how the simulations used for testing relate to the samples used in the training since particular initialisations of the tests might involve states and behaviours not present in the training data.
In the example given in Fig.~\ref{fig:maze_fig}, it is possible to understand the looping behaviour from a manual inspection alone.
However, this may be impossible in environments with a high number of states, such as continuous environments or for higher dimensions.
Moreover, other considerations are needed to understand the evaluation results of IL agents, such as poor performance due to drastic state variance from training and testing, or good performance due to lower variance and \textit{data leakage}, which occurs when the training data is present in the test set.

%
Secondly, with the popularity increase of ML approaches, benchmarking datasets have helped to evaluate how different methodologies perform under the same conditions and some applications hide test data from researchers to maintain evaluation integrity~\cite{everingham2015pascal,olga2015imagenet}.
Moreover, using the same dataset (and the same proportion of training/testing split) allows researchers to validate each other's results more efficiently, and simplifies the evaluation process because there is no need to run every relevant baseline over the data again.
RL work uses the same benchmarking approach, with the performance of RL methods evaluated on popular common environments~\cite{raffin2018rlzoo}.
Unfortunately, the same uniformity in evaluation has not yet been employed in IL methods, which usually create customised training data and compare performance with baselines over the data for which the other methods were not initially created.
Therefore, when comparing a new approach, a teacher dataset must be created, available reliable code found (with hyperparameters or requiring a vast search if wanting to compare with the best results for each method), and every baseline must be run over the generated data, leaving much room for issues, such as lack of data or diversity.
This divergence in benchmarking further corroborates the main problems from the disparities in the test process~\cite{zheng2021imitation}.

Finally, we argue that IL's ultimate goal is to learn how to accomplish a task using the teacher's data and not simply to copy the actions taken by the teacher.
In other words, the IL agent should act as the teacher would in previously seen states, but it should also somehow learn how the teacher would act in unseen states. 
For example, in Fig.~\ref{fig:maze_all}, given state $s_x$, `up' and `right' are optimal actions \textit{ideally} to be taken by the IL agent.
If that is not possible, the agent should at least attempt to choose an action that takes it back to an already-seen trajectory.
Analogously, the problem highlighted in Fig.~\ref{fig:maze_some} stems from the fact that the IL alternates between competing trajectories without ever completing either.

\subsection{Expert optimality feasibility} \label{sec:expert}

Imitation learning always assumes that training samples come from the execution of a task by an {\em expert}~\cite{hussein2017imitation}. 
While it is fair to assume that the source is proficient, the implicit presumption of the \textit{theoretical} optimality of the demonstration itself is problematic. 
Firstly, the more complex the environment, the harder it is to ascertain whether the samples are optimal. 
For example, while it might be easy to identify which episodes are near optimality in the CartPole environment~\cite{barto1983neuronlike} (where performance is considered optimal when the accumulated rewards match the number of steps), it is much more challenging to evaluate how `good' an agent is in autonomous driving scenarios, such as SUMO~\cite{lopez2018sumo}. 
In particular, some important aspects of the driver's behaviour (e.g., \textit{situational awareness}) may be difficult to fully measure.
Moreover, in highly dynamic systems such as in real-world driving, how can one be sure that the samples collected fully capture all aspects of the optimality of the execution?
In practice, even though the teacher might be \textit{proficient}, the samples collected may be corrupt or not cover all the experiences necessary for the learner to achieve the same level of optimality. 
Although it is possible to evaluate performance in both cases, it might be impractical to do so. 
Since reproducibility is crucial, we expect an increasing focus on presenting more data about the teacher's behaviour, an aspect that is lacking in most work, except for~\cite{gavenski2022how,belkhale2023data}.

Secondly, we argue that the demonstrations may not necessarily fully capture the intended optimal behaviour. For example, they may not necessarily achieve other secondary goals, such as fairness and safety, that may not play a part in the scenarios available for the learning. 
We note that safe and fair agents are not inherently suboptimal, but as in the first scenario, it might be impossible to create a perfect fitness function to account for the desired behaviour.
In these cases, samples may unintentionally be used where the teacher does not fully constrain actions, resulting in less than theoretical optimal agents (with respect to e.g., safety requirements).
For example, suppose an agent is trained to drive a car and uses samples from a human driver, who may drive incautiously.
In this case, the agent will learn like its teacher, which is optimal for some driving scenarios but not optimal in situations such as emergencies where the driver must act with extreme caution.
We therefore believe that only striving for optimal agents is not always sufficient.

Furthermore, consideration for aspects beyond the agent's overall reward with 
a focus on human feedback is central to `Reinforcement Learning from Human Feedback'~\cite{christiano2017rlhf,fernandes2023bridging}. 
Here, theoretical optimality is sacrificed for other four aspects.
Firstly, humans can be more nuanced than machines when taking action and safety and ethical considerations override overall performance. 
Secondly, humans have intrinsic notions of solution quality that are difficult to capture in the design of reward functions. 
Human feedback reduces the need for intricate designs.
Thirdly, human guidance in the agent's exploration phase helps to reduce the training time.
Lastly, human feedback allows for \textit{curriculum learning}, where the agent starts with human-curated samples and gradually transitions to more autonomous learning paradigms.
Since imitation learning, by definition, relies on human examples
to guide the entire learning process, we argue that IL agents should also consider these qualitative aspects, which are possible avenues for future research.

Therefore, the trade-off between expert and teacher agents has to be understood when developing a new agent.
In some cases, such as robotics for manufacturing, it is desirable to employ expert policies because they are guaranteed to be most efficient. 
In others, such as driving, optimal behaviour may be difficult to demonstrate, and teacher policies are acceptable.
We argue that by understanding the role of experts and teachers in a task, researchers can make more informative decisions about the most appropriate type of
demonstrations and what the performance evaluation really measures.

\begin{figure}[h]
    \centering
    \begin{minipage}{0.5\textwidth}
        \centering
        \begin{tikzpicture}[
    textual/.style={
        font=\bfseries,
        align=center,
        anchor=center,
    },
    node distance=3cm,
]
    \draw (0, 0) rectangle (10, 5);
    \draw (5, 0) -- (5, 5);
    \draw (0, 2.5) -- (5, 2.5);

    \node [textual] at (2.5, 4.7) {Non-optimal policies $\Nonoptimal$};  
    \node [textual] at (2.5, 0.2) {Optimal policies $\Optimal$};  
    \node [textual] at (2.5, -0.3) {Solutions};
    \node [textual] at (7.5, -0.3) {Not solutions};
    \node [textual] at (5, 5.3) {All policies $\Policy$};

    \node [at={(3.2, 3.7)}, align=center] {$\Teacher$\\Teacher policies};
    \draw [color=black] (2.5, 2.5) ellipse [x radius = 1.5, y radius = 2.5, rotate = -55];
    
    \node [at={(1.8, 1.6)}, align=center] {$\Expert$\\Expert policies};
    \draw [color=black] (2, 1.6) ellipse [x radius = 0.7, y radius = 1.5, rotate = -75];
\end{tikzpicture}
        \caption{Diagram of teacher, optimal and expert policy sets.}
        \Description{Diagram of teacher, optimal and expert policy sets.}
        \label{fig:challenges}
    \end{minipage}
    \hfill
    \begin{minipage}{0.45\textwidth}
        \footnotesize
        \centering
        \captionof{table}{Necessary requirements for expert validation.}
        \label{tab:conditions}
        \begin{tabular*}{\textwidth}{l@{\extracolsep{\fill}}ll}
            \toprule
            \multicolumn{2}{l}{\textbf{Aspect}} & \textbf{Description} \\ \midrule
            (i)   & State space           & Finite / Discrete                \\ \midrule
            (ii)  & Action space          & Finite / Discrete                \\ \midrule
            (iii) & State observability   & Fully observable                 \\ \midrule
            (iv)   & MDP sample coverage          & All MDP states     \\
           \bottomrule
        \end{tabular*}
    \end{minipage}
\end{figure}

\subsection{Expert Necessary Requirements}

As we saw in the previous section, it is desirable and perhaps inevitable that IL agents sometimes learn from non-optimal sources.
If an environment follows an MDP (as described in Section~\ref{sec:background}), it has a universe of all possible policies $\Pi$ (that either succeed or fail to solve the Markovian problem).
Some successful policies are also optimal ($\Optimal$).
The training data does not usually contain all optimal policies, restricting itself to a subset of these (the `expert' policies $\Expert$).
In this scenario, researchers should strive to acquire all optimal policies for training the agent, such that the expert policy set is \textit{complete} ($\Expert=\Optimal$).
In general, it is impossible or impractical to do so, resulting in incomplete coverage.  
Furthermore, in practice, the successful policies used to teach a task or skill also contain policies that are not optimal (the `teacher' policies $\Teacher$, which we also call \textit{complete} when $\Teacher = \Optimal\cup \Nonoptimal$). Including non-optimal policies in $\Teacher$ can happen because researchers sometimes provide their own demonstrations, and it may not be possible to check them for optimality. Therefore, we argue that for clarity, we should call the policies used for training simply as \textit{teachers}, reserving the term `experts' for policies whose optimality has been ascertained. Fig.~\ref{fig:challenges} illustrates the relationship between all policies.

We now look at the role played by the teachers' policies.
Table~\ref{tab:conditions} presents a set of necessary requirements all of which need to be fulfilled for the source data to be reliably considered an expert.
(i) and (ii) relate to the state and action space from environments.
Most often, work that use continuous (infinite) states and actions spaces in their problem formulation impose strict assumptions, such as injective MDPs~\cite{zhu2020opolo} and trajectories with the same initial states~\cite{kidambi2021mobile}. 
These assumptions are not applicable in most practical scenarios or require extensive modifications.
Therefore, we can assume that work whose environments have infinite state and action spaces cannot guarantee optimality of the training data.
(iii) refers to the observability of the environments. Optimality of policies for environments with partial or faulty observations~\cite{yu2020giril} cannot be guaranteed.
(iv) considers how much of the state space the samples collected cover.
We argue that in cases where the samples do not fully cover the Markovian problem, we cannot guarantee that the agent fully learns the expert's behaviour.
Hence, the output from training might be an optimal policy but does not guarantee optimal behaviour in unseen states.
Nevertheless, we observe that although researchers might be able to fulfil these conditions for optimality, they should not be used as the sole criteria for the evaluation of the overall performance  of the learning approach, since they
are too restrictive. In real-world applications, it is often the case that one
of more conditions need to be relaxed for practical use.

\subsection{Challenges and Future Directions} \label{sec:challenges}

Throughout this survey, we have discussed several challenges that imitation learning agents face.
This section highlights some of them and discusses possible future directions for imitation learning research.

\noindent
\textbf{Safety} is a crucial aspect of agent-based systems.
Without it, agents can cause harm to themselves, other agents, or the environment.
However, for imitation learning agents, safety can be hard to achieve since they learn through a teacher's perspective and without a direct signal.
When learning via observation, the samples might lack information regarding the environment's dynamics, leading to unsafe behaviour due to the agent's lack of knowledge.
Moreover, the teacher might not behave safely, leading to an unsafe agent.
Thus, we believe there is space to research how imitation learning agents can learn safely from teachers.
Possibly, imitation learning agents should learn how to differentiate between safe and unsafe samples and teachers.

\noindent
\textbf{Learning efficiency} became a trend for imitation learning algorithms~\cite{monteiro2023self,yin2022planning,torabi2021dealio,wen2019efficient}.
For imitation learning, researchers measure efficiency by the number of samples needed to learn a task.
However, we believe that efficiency by itself might be a misleading metric.
For example, an agent who learns a task by only observing a single episode from a teacher might be considered efficient.
Conversely, suppose the agents fail to generalise to other scenarios or act unsafely. 
In that case, their efficiency is not as good as it seems.
Thus, efficiency should be measured by the number of samples needed to learn a task while measuring other factors, such as safety.

\noindent
\textbf{Learning effectiveness} is often overlooked by imitation learning researchers.
In this work, we define effectiveness as the ability of an agent to learn a task or skill optimally (or close to it) with non-optimal teachers.
Effectiveness is as crucial as efficiency since it is not enough to learn a task or skill with few samples if the agent does not learn it well.
Moreover, we believe effectiveness is more challenging than efficiency due to how imitation learning agents learn.
Considering our premise that is impossible to know how optimal a teacher is in real-world applications, it is essential to develop agents that can learn effectively from non-optimal teachers.
Thus, we believe that imitation learning researchers should focus more on effectiveness than efficiency.  

\noindent
\textbf{Adaptability} comes from combining generalisation and \textit{transfer learning}.
Generalisation is the ability of an agent to learn a task or skill from a teacher and apply it to other unseen scenarios (similar to effective learning).
Transfer learning is the ability of an agent to learn a task or skill from a teacher and apply it to other tasks or skills.
More formally, transfer learning in imitation learning refers to the agent's ability to translate knowledge it learned from one Markov Decision Process $M$ to another $M'$, either regarding the transition function $T$ (from $T$ to $T'$) or in the form of a new immediate reward function $r$.
Considering that imitation learning agents do not have access to $r$, this form of transfer learning usually appears in the sense of $T$ remaining the same but the goal of the environment changing.
Given the nature of the imitation learning agent's training and lack of a direct signal, we believe these two aspects are essential to strive for.
By achieving generalisation or transfer learning, imitation learning agents can succeed more in real-world scenarios.
However, researchers should focus on learning paradigms outside imitation learning, such as symbolic learning, to achieve these goals. 

\noindent
\textbf{Multi-agent systems} are not a research topic as common as single agents for imitation learning researchers.
We believe such behaviour comes from imitation learning being more adequate for learning low-level tasks, which usually multi-agent systems do not have.
However, imitation learning can be used to learn high-level tasks, such as coordination and communication, which are essential for multi-agent systems.
Imitation learning is a highly adaptable process that can be employed with other techniques, such as planning~\cite{yin2022planning,pulver2021pilot}, explorations~\cite{kidambi2021mobile,gavenski2020iupe}.
Perhaps using imitation learning on more descriptive processes, such as `Belief Desire and Intention', can help imitation learning agents achieve better results on multi-agent systems scenarios.

\section{Conclusions} \label{sec:conclusion}

In this survey, we present a comprehensive review of imitation learning methods, environments and metrics. 
We provided background knowledge required to understand the field and how different imitation learning methods use data to learn a policy.
One of our contributions is the provision of novel taxonomies for methods, environments and metrics.

We first classified {\em methods} according to new trends in the field, highlighting key aspects. This should help researchers identify new opportunities without supplanting the traditional model-free/model-based dichotomy.

Our new taxonomy for \emph{environments} focuses on how they are used in the evaluation process, i.e., for \textit{validation}, to assess \textit{precision}, or whether they are \textit{sequential} -- when some actions can influence the choice of future actions. This focus on purpose should help proponents of new methods choose the most appropriate environment for their methods, allowing for more effective evaluations.

We then systematically reviewed the \textit{metrics} used to evaluate imitation learning methods, partitioning them into the three categories \textit{behaviour}, \textit{domain}, and \textit{model}, which can be further refined into qualitative and quantitative. 
We highlighted the importance of these categories in the evaluation of the corresponding characteristics of the method's performance. 
Given the variety of metrics, an open question is whether and how they can be aggregated to rank or better compare methods more generally.

Finally, we showed a lack of standardisation in the evaluation process of the imitation learning field, making it difficult to compare methods uniformly under specific perspectives.
To the best of our knowledge, this is the first IL survey to comprehensively and systematically review IL methods, environments and metrics in this way.

While there has been much recent progress, we believe the imitation learning field still shows much potential for future research. 
This survey provides a solid foundation from which to further develop the area -- one that allows the systematic evaluation of new approaches with regard to the state-of-the-art. 
In particular, by contrasting the different ways in which existing work has been evaluated, we see that there is much to improve in terms of the methods' resulting behaviour beyond the use of the environmental metrics that traditional reinforcement learning employs. 
More specifically, imitation learning can benefit from work in other research fields that strive for more human-like behaviour, such as reinforcement learning from human feedback.


\bibliographystyle{ACM-Reference-Format}
\bibliography{bibliography}

\appendix
\section{Validation environments}
\tiny
\begin{xltabular}{\textwidth}{l@{\extracolsep{\fill}}llllll}
\caption{%
    Validation environments information. 
    Environments from benchmarks are grouped by their respective suite.%
} \label{tab:env_validation} \\

\toprule Environments & Domain Type & Action & State & Deterministic & Accessibility & Dynamics \\ \midrule
\endfirsthead

\toprule Environments & Domain Type & Action & State & Deterministic & Accessibility & Dynamics \\ \midrule
\endhead

\midrule
\endfoot

\bottomrule
\endlastfoot

Bandits~\cite{shih2022conditional} & Task specific & discrete & discrete & deterministic & partial observable & static \\
Hanabi~\cite{bard2020hanabi} & Task specific & discrete & discrete & deterministic & partial observable & dynamic \\
RoboCup~\cite{kitano1997robocup} & Task specific & continuous & continuous & deterministic & fully observable & dynamic \\
Sort~\cite{chernova2008teaching} & Task specific & discrete & continuous (Image) & deterministic & fully observable & dynamic \\
Predator-Prey,
Keep-Away\cite{lowe2017multi},
Particle\cite{mordatch2017emergence} & Task specific & continuous & discrete & deterministic & fully observable & dynamic \\
TurtleBot~\cite{pathak2018zeroshot},
2D Plane~\cite{yunzhu2017infogail} & Domain specific & continuous & continuous & deterministic & fully observable & static \\ \midrule

\multicolumn{7}{c}{\textbf{Classic control environments}} \\ \midrule
Acrobot~\cite{sutton1995generalization},
Pendulum                                & Task specific & continuous & continuous & deterministic & fully observable & static \\ 
CartPole~\cite{barto1983neuronlike},
MountainCar~\cite{moore1990efficient}   & Task specific & discrete & continuous & deterministic & fully observable & static  \\\midrule

\multicolumn{7}{c}{\textbf{MuJoCo environments}} \\ \midrule
Ant~\cite{schulman2015high},
HalfCheetah~\cite{wawrzynski2009cat},              
Hopper~\cite{tom2011infinite}, & \multirow{4}{*}{Task specific} & \multirow{4}{*}{continuous} & \multirow{4}{*}{continuous} & \multirow{4}{*}{deterministic} & \multirow{4}{*}{fully observable} & \multirow{4}{*}{static} \\
Humanoid~\cite{tassa2012synthesis},
InvertedPendulum~\cite{barto1983neuronlike}, \\
InvertedDoublePendulum~\cite{barto1983neuronlike},
Pusher, \\
Reacher,
Swimmer~\cite{coulom2002reinforcement},
Walker2D~\cite{tom2011infinite} \\ \midrule

\multicolumn{7}{c}{\textbf{DeepMind Control Suite environments}~\cite{tassa2018deepmind}} \\ \midrule
Ball-in-cup,  
Cheetah Run, 
Finger Spin,   & \multirow{3}{*}{Task specific} & \multirow{3}{*}{continuous} & \multirow{3}{*}{continuous} & \multirow{3}{*}{deterministic} & \multirow{3}{*}{fully observable} & \multirow{3}{*}{static} \\          
Hopper Hop,         
Humanoid Walk, \\  
Swingup,        
Reacher Easy,            
Walker Walk \\ \midrule

\multicolumn{7}{c}{\textbf{Atari environments}~\cite{bellemare2013atari}} \\ \midrule
Alien, 
Beam Rider,              
Breakout,
Taxi, & \multirow{4}{*}{Task specific} & \multirow{4}{*}{discrete} & \multirow{4}{*}{both (Image or RAM)} & \multirow{4}{*}{deterministic} & \multirow{4}{*}{fully observable} & \multirow{4}{*}{static}  \\
Kung-Fu Master,                
MsPackman,             
PitFall, \\
Pong,             
Private Eye,
Q*bert,            
Seaquest, \\
Space Invaders,
Montezuma \\

\end{xltabular}

\section{Precision environments}
\begin{table}[h]
    \tiny
    \centering
    \caption{Precision environments information. Environments from benchmarks are grouped by their respective suite.}
    \label{tab:env_precision}
    \begin{tabular*}{\textwidth}{l@{\extracolsep{\fill}}llllll}
        \toprule
        Environments & Domain Type & Action & State & Deterministic & Accessibility & Dynamics \\ 
        \midrule
        Jaco~\cite{pierre2020jaco} & Domain specific & continuous & continuous (Image and Vector) & deterministic & fully observable & static \\
        MAV~\cite{daftry2017mav} & Domain specific & continuous & continuous & deterministic & fully observable & static \\ 
        Panda~\cite{gallouedec2021panda} & Both & continuous & continuous (Image and Vector) & deterministic & fully observable & static \\ \midrule
        \multicolumn{7}{c}{\textbf{MuJoCo environments}~\cite{todorov2012mujoco}} \\ \midrule
        Disc,
        DoorOpening,
        PegInserton, & \multirow{2}{*}{Domain specific} & \multirow{2}{*}{continuous} & \multirow{2}{*}{continuous} & \multirow{2}{*}{deterministic} & \multirow{2}{*}{fully observable} & \multirow{2}{*}{static} \\
        GripperPusher,
        Point \\
        Robot body,
        Robotic Arm   & Both            & continuous & continuous & deterministic & fully observable & static \\
        \bottomrule
    \end{tabular*}
    \end{table}

\section{Sequential environments}
\begin{table}[h!]
\tiny
\centering
\caption{Sequential environments information. RAM means that states are vector representations of values stored in memory.}
\label{tab:env_sequential}
\begin{tabular*}{\textwidth}{l@{\extracolsep{\fill}}llllll}
    \toprule
    Environments & Domain Type & Action   & State & Deterministic & Accessibility & Dynamics \\ 
    \midrule
    Maze~\cite{matthew2016maze}             & Task specific & discrete & Discrete (Image or Vector) & deterministic & fully observable & static \\
    CoinRun~\cite{cobbe2019coinrun}         & Task specific & discrete & continuous (Image) & both & both & dynamic \\
    TORCS~\cite{wymann2014torcs}            & Domain specific & both & continuous (Image and RAM)  & deterministic & fully observable & dynamic \\
    VizDoom~\cite{Kempka2016ViZDoom}        & Domain specific &  discrete & continuous (Image) & deterministic & fully observable  & dynamic \\
    Blocks World~\cite{yan2017oneshot}      & Domain specific & continuous & continuous (Image)  & deterministic & fully observable & static   \\
    LunarLander                             & Task specific & discrete & continuous & deterministic & fully observable & static \\
    CARLA~\cite{dosovitskiy17carla},
    SUMO~\cite{lopez2018sumo}               & Domain specific & continuous & continuous (Image or Vector) & deterministic & fully observable & dynamic \\
    Mario~\cite{kauten2018mario},            
    Minecraft~\cite{johnson2016malmo},      & \multirow{2}{*}{Task specific} & \multirow{2}{*}{discrete} & \multirow{2}{*}{continuous (Image or RAM)} & \multirow{2}{*}{deterministic} & \multirow{2}{*}{fully observable} & \multirow{2}{*}{dynamic} \\
    Overcooked~\cite{carroll2019overcooked},
    SuperTuxKart~\cite{henrichs2006stk} \\
    \bottomrule
\end{tabular*}
\end{table}

\end{document}